\documentclass{article}

\makeatletter
\def\@seccntformat#1{\@ifundefined{#1@cntformat}%
   {\csname the#1\endcsname\space}%    default
   {\csname #1@cntformat\endcsname}}%  enable individual control
\newcommand\section@cntformat{\thesection.\quad}       % section-level
\newcommand\subsection@cntformat{\thesubsection.\quad} % subsection-level
\newcommand\subsubsection@cntformat{\thesubsubsection.\quad} % subsection-level
\makeatother

\usepackage{arxiv}

\usepackage[utf8]{inputenc} % allow utf-8 input
\usepackage[T1]{fontenc}    % use 8-bit T1 fonts
\usepackage{hyperref}       % hyperlinks
\usepackage{url}            % simple URL typesetting
\usepackage{booktabs}       % professional-quality tables
\usepackage{amsfonts}       % blackboard math symbols
\usepackage{nicefrac}       % compact symbols for 1/2, etc.
\usepackage{microtype}      % microtypography
\usepackage{amsmath}
\usepackage{amsthm}
\usepackage{array}
\usepackage{booktabs}
\usepackage{eso-pic,graphicx}
\usepackage{float}
\usepackage{caption}
\usepackage{subcaption}
\usepackage{color}
\usepackage{url}
\usepackage{ulem}
\usepackage{epstopdf}
\usepackage{comment}
\usepackage{multirow}
\usepackage{algorithm,algorithmic}

\newcolumntype{L}[1]{>{\raggedright\let\newline\\\arraybackslash\hspace{0pt}}m{#1}}
\newcolumntype{C}[1]{>{\centering\let\newline\\\arraybackslash\hspace{0pt}}m{#1}}
\newcolumntype{R}[1]{>{\raggedleft\let\newline\\\arraybackslash\hspace{0pt}}m{#1}}
\newcommand{\Keywords}[1]{\textbf{\textit{Index terms --}} #1}
\newcommand{\sign}[1]{\text{sign}\left(#1\right)}
\newcommand{\parderiv}[2]{\frac{\partial #1}{\partial #2}}
\newcommand{\norm}[1]{\left\| #1 \right\|}

\title{Learning without feedback:\\ Fixed random learning signals\\allow for feedforward training\\of deep neural networks}

\author{
	Charlotte~Frenkel\thanks{These authors contributed equally.}\hspace{4pt}\thanks{C. Frenkel was with Universit\'e catholique de Louvain as a Research Fellow from the National Foundation for Scientific Research (FNRS) of Belgium. She is now with the Institute of Neuroinformatics, University of Z\"urich and ETH Z\"urich, Switzerland.}\\
  	ICTEAM Institute\\
  	Universit\'e catholique de Louvain\\
	Louvain-la-Neuve BE-1348, Belgium\\
	\texttt{charlotte@ini.uzh.ch}\\
\And
	Martin~Lefebvre\footnotemark[1]\\
  	ICTEAM Institute\\
	Universit\'e catholique de Louvain\\
	Louvain-la-Neuve BE-1348, Belgium\\
	\texttt{martin.lefebvre@uclouvain.be}\\
\And
	David~Bol\\
  	ICTEAM Institute\\
	Universit\'e catholique de Louvain\\
	Louvain-la-Neuve BE-1348, Belgium\\
	\texttt{david.bol@uclouvain.be}\\
	~\\
}

\definecolor{myred}{RGB}{247, 183, 163}

\AddToShipoutPicture*{\small \raisebox{26.3cm}
{\hspace*{2.2cm}
\noindent\fcolorbox{red}{myred}{%
\begin{minipage}[c]{1.02\textwidth}
{\setlength{\baselineskip}{0.5\baselineskip}
\normalsize
This paper has been accepted for publication in the \textit{Frontiers in Neuroscience} journal.\vspace*{1mm}\\
The fully-edited paper is available at \url{https://www.frontiersin.org/articles/10.3389/fnins.2021.629892},\\
with DOI 10.3389/fnins.2021.629892.
\par}
\end{minipage}}}}%

\begin{document}
\maketitle

\begin{abstract}
While the backpropagation of error algorithm enables deep neural network training, it implies (i) bidirectional synaptic weight transport and (ii) update locking until the forward and backward passes are completed. Not only do these constraints preclude biological plausibility, but they also hinder the development of low-cost adaptive smart sensors at the edge, as they severely constrain memory accesses and entail buffering overhead. In this work, we show that the one-hot-encoded labels provided in supervised classification problems, denoted as targets, can be viewed as a proxy for the error sign. Therefore, their fixed random projections enable a layerwise feedforward training of the hidden layers, thus solving the weight transport and update locking problems while relaxing the computational and memory requirements. Based on these observations, we propose the direct random target projection (DRTP) algorithm and demonstrate that it provides a tradeoff between accuracy and computational cost that is suitable for adaptive edge computing devices.
\end{abstract}

% keywords can be removed

\vspace*{3mm}
\section{Introduction}

Artificial neural networks (ANNs) were proposed as a first step toward bio-inspired computation by emulating the way the brain processes information with densely-interconnected neurons and synapses as computational and memory elements, respectively~\cite{Rosenblatt61,Bassett06}. In order to train ANNs, it is necessary to identify how much each neuron contributed to the output error, a problem referred to as the \textit{credit assignment}~\cite{Minsky61}. The backpropagation of error (BP) algorithm~\cite{Rumelhart86} allowed solving the credit assignment problem for multi-layer ANNs, thus enabling the development of deep networks for applications ranging from computer vision~\cite{Krizhevsky12,LeCun15,He16} to natural language processing~\cite{Hinton12,Amodei16}. However, two critical issues preclude BP from being biologically plausible.

\begin{figure}
	\centering
	\vspace*{-5mm}
	\includegraphics[width=0.95\textwidth]{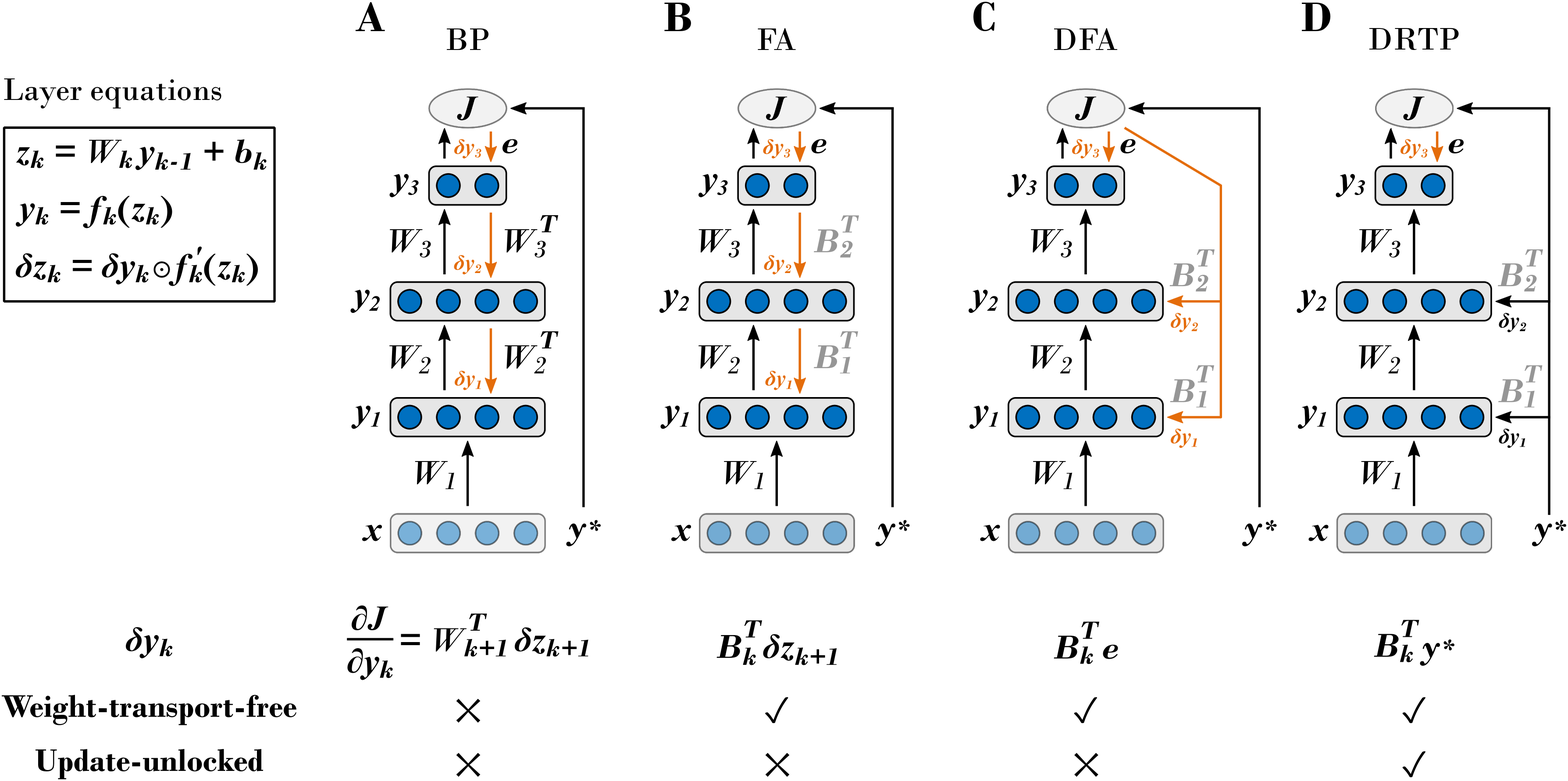}
	\caption{\textbf{The proposed direct random target projection algorithm builds on feedback-alignment-based algorithms to tackle the weight transport problem while further releasing update locking.} Black arrows indicate the feedforward pathways and orange arrows the feedback pathways. In the $k$-th layer, the weighted sum of inputs $y_{k-1}$ is denoted as $z_k$, the bias as $b_k$, the activation function as $f_k(\cdot)$ and its derivative as $f'_k(\cdot)$, with $k \in [1,K]$, $k \in \mathbb{N}$, and $K$ the number of layers. Trainable forward weight matrices are denoted as $W_k$ and fixed random connectivity matrices as $B_k$. The input vector is denoted as $x$, the target vector as $y^*$ and the loss function as $J(\cdot)$. The estimated loss gradients for the outputs of the $k$-th hidden layer, denoted as $\delta y_k$, are provided for each training algorithm. The layer equations for $z_k$, $y_k$ and $\delta z_k$, defined as the modulatory signals, are provided in the upper left corner, with $\odot$ denoting the elementwise multiplication operator. \textbf{(A)} Backpropagation of error (BP) algorithm~\cite{Rumelhart86}. \textbf{(B)} Feedback alignment (FA) algorithm~\cite{Lillicrap16}. \textbf{(C)} Direct feedback alignment (DFA) algorithm~\cite{Nokland16}. \textbf{(D)} Proposed direct random target projection (DRTP) algorithm. Adapted from \cite{Nokland16} and \cite{Czarnecki17}.}
	\label{fig:topologies}
\end{figure}

First, BP requires symmetry between the forward and backward weights, which is known as the \textit{weight transport problem}~\cite{Grossberg87}. Beyond implying a perfect and instantaneous communication of parameters between the feedforward and feedback pathways, error backpropagation requires each layer to have full knowledge of all the weights in the downstream layers, making BP a non-local algorithm for both weight and error information. From a hardware efficiency point of view, the weight symmetry requirement also severely constrains memory access patterns~\cite{Crafton19}. Therefore, there is an increasing interest in developing training algorithms that release this constraint, as it has been shown that weight symmetry is not mandatory to reach near-BP performance~\cite{Liao16}. The feedback alignment (FA) algorithm~\cite{Lillicrap16}, also called random backpropagation~\cite{Baldi18}, demonstrates that using fixed random weights in the feedback pathway allows conveying useful error gradient information: the network learns to align the forward weights with the backward ones. Direct feedback alignment (DFA)~\cite{Nokland16} builds on these results and directly propagates the error between the network predictions and the \textit{targets} (i.e.~one-hot-encoded labels) to each hidden layer through fixed random connectivity matrices. DFA demonstrates a limited accuracy penalty compared to BP on the MNIST~\cite{LeCun98} and CIFAR-10~\cite{Krizhevsky09} datasets, while using the output error as a global modulator and keeping weight information local. Therefore, DFA bears important structural similarity with learning rules that are believed to take place in the brain~\cite{Guerguiev17,Neftci17}, known as three-factor synaptic plasticity rules, which rely on local pre- and post-synaptic spike-based activity together with a global modulation~\cite{Urbanczik14}. Finally, another approach for solving the weight transport problem consists in computing targets for each layer instead of gradients. The target values can either be computed based on auto-encoders at each layer~\cite{Lee15} or generated by making use of the pre-activation of the current layer and the error of the next layer, propagated through a dedicated trainable feedback pathway~\cite{Ororbia19}. The BP, FA and DFA algorithms are summarized in Figures~\ref{fig:topologies}A--\ref{fig:topologies}C, respectively.

The second issue of BP is its requirement for a full forward pass before parameters can be updated during the backward pass, a phenomenon referred to as \textit{update locking}~\cite{Jaderberg17,Czarnecki17}. Beyond making BP biologically implausible, update locking has critical implications for BP implementation as it requires buffering all the layer inputs and activations during the forward and backward passes in order to compute the weight updates, leading to a high memory overhead. As the previously-described FA and DFA solutions to the weight transport problem only tackle the weight locality aspect, specific techniques enabling local error handling or gradient approximation are required to tackle update locking. On the one hand, the \textit{error locality approach} relies on layerwise loss functions~\cite{Mostafa18,Kaiser18,Nokland19,Belilovsky19}, it enables training layers independently and without requiring a forward pass in the entire network. The generation of local errors can be achieved with auxiliary fixed random classifiers, allowing for near-BP performance on the MNIST and CIFAR-10 datasets~\cite{Mostafa18}. This strategy has also been ported to a biologically-plausible spike-based three-factor synaptic plasticity rule~\cite{Kaiser18}. Scaling to ImageNet~\cite{Deng09} requires either the use of two combined layerwise loss functions~\cite{Nokland19} or a parallel optimization of a greedy objective using deeper auxiliary classifiers~\cite{Belilovsky19}. However, the error locality approach still suffers from update locking at the layer scale as layerwise forward and backward passes are required. Beyond implying a computational overhead, the auxiliary classifiers also suffer from the weight transport problem, a requirement that can only be partially relaxed: in order to maintain performance, it is necessary to keep at least the weight sign information during the layerwise backward passes~\cite{Mostafa18}. On the other hand, the \textit{synthetic gradients approach}~\cite{Jaderberg17,Czarnecki17} relies on layerwise predictors of subsequent network computation. However, training local gradient predictors still requires backpropagating gradient information from deeper layers.

In order to fully solve both the weight transport and the update locking problems, we propose the direct random target projection (DRTP) algorithm (Fig.~\ref{fig:topologies}D). Compared to DFA, the targets are used in place of the output error and projected onto the hidden layers. We demonstrate both theoretically and experimentally that, in the framework of classification problems, the error sign information contained in the targets is sufficient to maintain feedback alignment with the loss gradients $\delta z_k$ for the weighted sum of inputs in layer $k$, denoted as the \textit{modulatory signals} in the subsequent text, and allows training multi-layer networks, leading to three key advantages. First, DRTP solves the weight transport problem by entirely removing the need for dedicated feedback pathways. Second, layers can be updated independently and without update locking as a full forward pass is not required, thus reducing memory requirements by releasing the need to buffer inputs and activations of each layer. Third, DRTP is a purely feedforward and low-cost algorithm whose updates rely on layerwise information that is immediately available upon computation of the layer outputs. Estimating the layerwise loss gradients $\delta y_k$ only requires a label-dependent random vector selection, contrasting with the error locality and synthetic gradients approaches that require the addition of side networks for error or gradient prediction. DRTP even compares favorably to DFA, as the latter still requires a multiplication between the output error and a fixed random matrix.

Therefore, DRTP allows relaxing structural, memory and computational requirements, yet we demonstrate that, compared to BP, FA and DFA, DRTP is ideal for implementation in edge-computing devices, thus enabling adaptation to uncontrolled environments while meeting stringent power and resource constraints. Suitable applications for DRTP range from distributed smart sensor networks for the Internet-of-Things (IoT)~\cite{Bol15} to embedded systems and cognitive robotic agents~\cite{Milde17}. The MNIST and CIFAR-10 datasets have thus been selected for benchmarking as they are representative of the complexity level required in autonomous always-on adaptive edge computing, which is not the case of larger and more challenging datasets such as ImageNet. This furthermore highlights that edge computing is an ideal use case for biologically-motivated algorithms, as an out-of-the-box application of feedback-alignment- and target-propagation-based algorithms currently does not scale to complex datasets (see \cite{Bartunov18} for a recent review). We demonstrate this claim in \cite{Frenkel20} with the design of an event-driven convolutional processor that requires only 16.8-\% power and 11.8-\% silicon area overheads for on-chip online learning, a record-low overhead that is specifically enabled by DRTP, thus highlighting its low cost for edge computing devices. Finally, as DRTP can also be formulated as a three-factor learning rule for biologically-plausible learning, it is suitable for embedded neuromorphic computing, in which high-density synaptic plasticity can currently not be achieved without compromising learning performance~\cite{Frenkel19a,Frenkel19b}.

\section{Results} \label{sec_results}

\subsection{Weight updates based only on the error sign provide learning to multi-layer networks.}\label{ssec:sign_learning}

We demonstrate with two experiments, respectively on a regression task and a classification problem, that modulatory signals based only on the error sign are within 90$^\circ$ of those prescribed by BP, thus providing learning in multi-layer networks. To do so, we use an error-sign-based version of DFA, subsequently denoted as sDFA, in which the error vector is replaced by the error sign in the global feedback pathway.

\begin{figure}
	\centering
	\includegraphics[width=\textwidth]{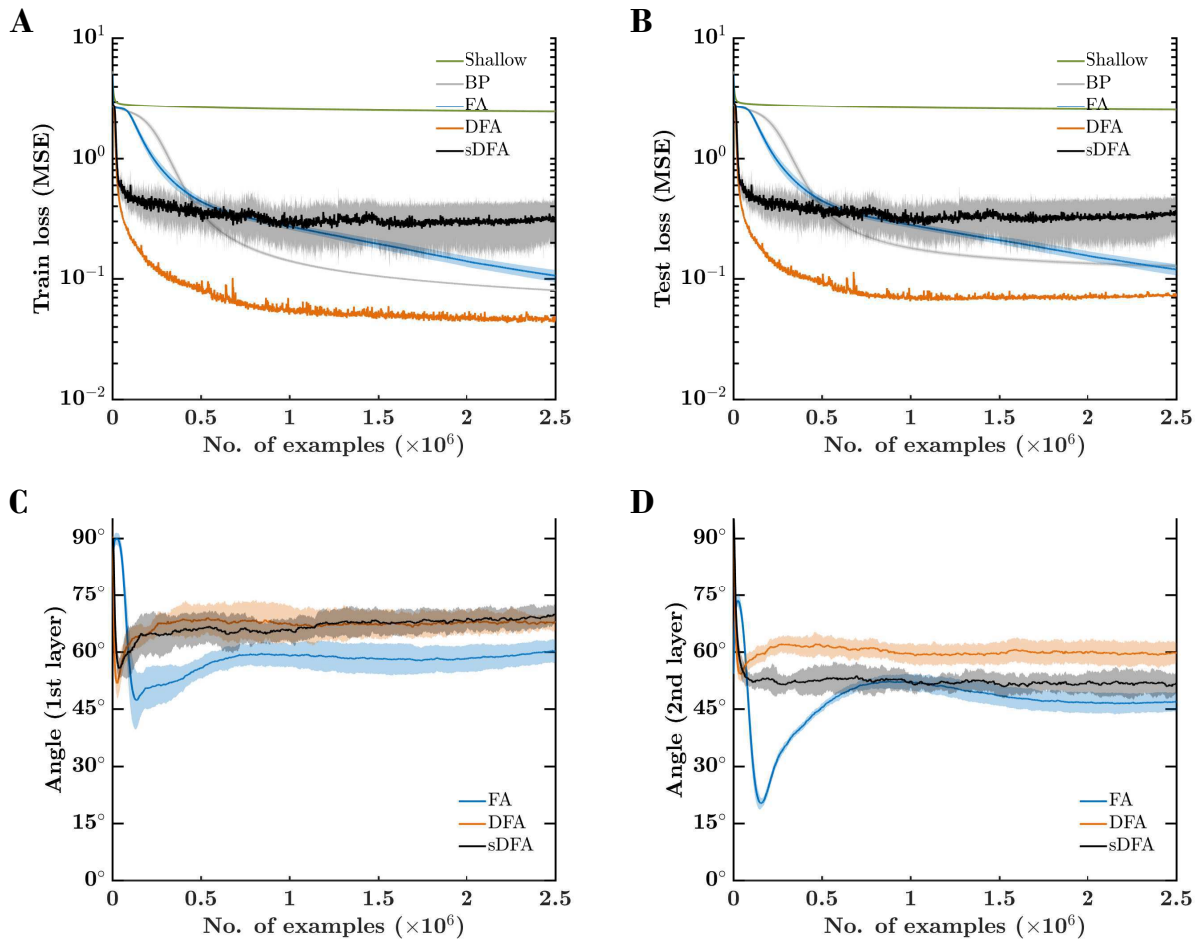}
	\caption{\textbf{Error-sign-based direct feedback alignment (sDFA) provides useful modulatory signals in regression tasks.} A 256-100-100-10 network with tanh hidden and output units is trained to learn cosine functions with five training algorithms: shallow learning, BP, FA, DFA and sDFA. With this simple setup, BP and FA suffer from the vanishing gradients problem, which would be alleviated by using ReLU-based networks with batch normalization. The scope of the figure is to highlight that sDFA provides useful modulatory signals for regression tasks, without any additional technique. As for other feedback-alignment-based algorithms, sDFA updates are within 90$^\circ$ of the backpropagation updates. The train and test losses and the alignment angles are monitored every 1k samples, error bars are one standard deviation over 10 runs. Angles have been smoothed by an exponentially-weighted moving average filter with a momentum coefficient of 0.95. \textbf{(A)} Mean squared error loss on the 5k-example training set. \textbf{(B)} Mean squared error loss on the 1k-example test set. \textbf{(C)} Angle between the modulatory signals $\delta z_k$ prescribed by BP and by feedback-alignment-based algorithms in the first hidden layer. \textbf{(D)} Angle between the modulatory signals $\delta z_k$ prescribed by BP and by feedback-alignment-based algorithms in the second hidden layer.}
	\label{fig:regression}
\end{figure}

\subsubsection{Regression} \label{ssec_regression}

This first experiment aims at demonstrating that the error sign provides useful modulatory signals to multi-layer networks by comparing training algorithms on a regression task. The objective is to approximate 10 nonlinear functions $T_j(x) = \text{cos}(\overline{x}+\phi_j)$, where $\phi_j = -\pi/2 + j\pi/9$ for $j \in [0,9]$, $j \in \mathbb{N}_0$ and $\overline{x}$ denotes the mean of $x$, a 256-dimensional vector whose entries are drawn from a normal distribution with a mean lying in $[-\pi,\pi]$~(see Section~\ref{sec_methods}). A 256-100-100-10 fully-connected network is trained to approximate $T(\cdot)$ with five training algorithms: shallow learning (i.e.~frozen random hidden layers and a trained output layer), BP, FA, DFA and sDFA. 

The mean squared error (MSE) loss on the training set is shown in Figure~\ref{fig:regression}A. While shallow learning fails to learn a meaningful approximation of $T(\cdot)$, sDFA and DFA show the fastest initial convergence due to the separate direct feedback pathway precluding gradients from vanishing, which is clearly an issue for BP and FA. Although this would be alleviated by using ReLU-based networks with batch normalization~\cite{Ioffe15}, it highlights that direct-feedback-alignment-based methods do not need further techniques such as batch normalization to address this issue, ultimately leading to reduced hardware requirements. While DFA demonstrates the highest performance on this task, sDFA comes earlier to stagnation as it does not account for the output error magnitude reduction as training progresses, thus preventing a reduction of the effective learning rate in the hidden layers as the output error decreases. sDFA could therefore benefit from the use of a learning rate scheduler. Similar conclusions hold for the loss on the test set (Figure~\ref{fig:regression}B). The angle between the modulatory signals prescribed by BP and by feedback-alignment-based algorithms is shown in Figures~\ref{fig:regression}C and~\ref{fig:regression}D for the first and second hidden layers, respectively. While all feedback-alignment-based algorithms lie close to each other within 90$^\circ$ of the BP modulatory signals, FA has a clear advantage during the first 100 epochs on the 5k-example training set. sDFA performs on par with DFA in the first hidden layer, while it surprisingly provides a better alignment than DFA in the second hidden layer, though not fully leveraged due to the absence of modulation in the magnitude of the updates from the output error.

\begin{figure}
	\centering
	\includegraphics[width=\textwidth]{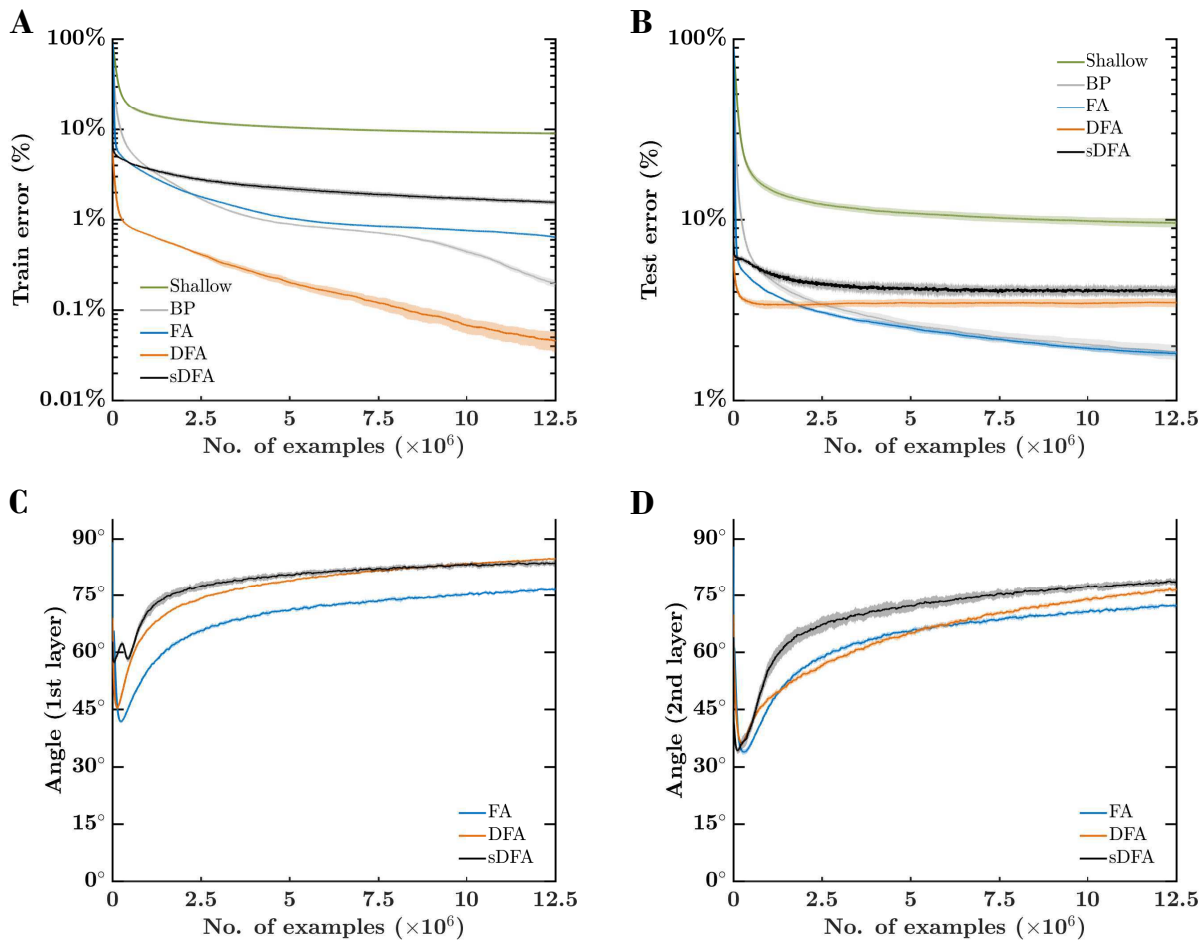}
	\caption{\textbf{Error-sign-based direct feedback alignment (sDFA) provides useful modulatory signals in classification tasks.} A 256-500-500-10 network with tanh hidden units and sigmoid output units is trained to classify a synthetic dataset of 16$\times$16-pixel images into 10 classes with five training algorithms: shallow learning, BP, FA, DFA and sDFA. With this simple setup, BP and FA suffer from the vanishing gradients problem, which would be alleviated by using ReLU-based networks with batch normalization. The scope of the figure is to highlight that sDFA provides useful modulatory signals for classification tasks, without any additional technique. The update directions of the sDFA algorithm are within 90$^\circ$ of the backpropagation updates and are comparable to other feedback-alignment-based algorithms. The train and test losses and the alignment angles are monitored every 2.5k samples, error bars are one standard deviation over 10 runs. Angles have been smoothed by an exponentially-weighted moving average filter with a momentum coefficient of 0.95. \textbf{(A)} Error on the 25k-example training set, reaching on average 0.19\% for BP, 0.64\% for FA, 0.05\% for DFA, 1.54\% for sDFA and 8.95\% for shallow learning after 500 epochs. \textbf{(B)} Error on the test set, reaching on average 1.85\% for BP, 1.81\% for FA, 3.48\% for DFA, 4.07\% for sDFA and 9.57\% for shallow learning after 500 epochs. \textbf{(C)} Angle between the modulatory signals $\delta z_k$ prescribed by BP and by feedback-alignment-based algorithms in the first hidden layer. \textbf{(D)} Angle between the modulatory signals $\delta z_k$ prescribed by BP and by feedback-alignment-based algorithms in the second hidden layer.}
	\label{fig:classification}
\end{figure}

\subsubsection{Classification}\label{ssec_classification}

With this second experiment, we demonstrate that, in addition to providing useful modulatory signals for regression problems, the error sign information allows training multi-layer networks to solve classification problems. The task consists in training a 256-500-500-10 network to solve a synthetic classification problem with 16$\times$16-pixel images and 10 classes; the data to classify is generated automatically with the Python \texttt{sklearn} library~\cite{Pedregosa11} (see Section~\ref{sec_methods}). As for regression, the network is trained with shallow learning, BP, FA, DFA and sDFA.

Figure~\ref{fig:classification}A shows that, after 500 epochs with a 25k-example training set, DFA provides the fastest and most accurate training with a classification error of 0.05\%, followed by BP, FA and sDFA with 0.19\%, 0.64\% and 1.54\%, respectively. Shallow learning lags almost an order of magnitude behind with 8.95\%. However, Figure~\ref{fig:classification}B shows that DFA also has a higher overfitting and lies close to sDFA on the test set, with 3.48\% and 4.07\%, respectively. The lowest classification errors are of 1.85\% for BP and 1.81\% for FA, while shallow learning lags behind at 9.57\%. The angle between the modulatory signals prescribed by BP and by feedback-alignment-based algorithms is shown in Figures~\ref{fig:classification}C and~\ref{fig:classification}D, for the first and second hidden layers, respectively. As for the regression task, all feedback-alignment-based algorithms exhibit alignments close to each other, while the convergence of BP and FA is slowed down by the vanishing gradients problem. Here, alignments tend to level off after 50 epochs, with the lowest angle provided by FA, followed by DFA and sDFA. As sDFA is always within 90$^\circ$ of the BP modulatory signals, it is able to train multi-layer networks.

\vspace*{2mm}
\subsection{For classification, a feedback pathway is no longer required as the error sign is known in advance} \label{ssec:sign_known_advance}

In the framework of classification problems, training examples ($x$,$\:c^*$) consist of an input data sample to classify, denoted as $x$, and a label $c^*$ denoting the class $x$ belongs to, among $C$ possible classes. The target vector, denoted as $y^*$, corresponds to the one-hot-encoded class label $c^*$. The output layer nonlinearity must be chosen as a sigmoid or a softmax function, yielding output values that are strictly bounded between 0 and 1. Denoting the output vector of a $K$-layer network as $y_K$, the error vector is defined as $e = y^* - y_K$. Under the aforementioned conditions, it results that the $c$-th entry of the $C$-dimensional error vector $e$, denoted $e_c$, is defined as
\begin{align*}
	e_c =
	\left \lbrace
	\begin{aligned}
		&1-y_{Kc}	&&\quad \text{if} \quad c=c^*,\\
		&-y_{Kc} 	&&\quad \text{otherwise.}
	\end{aligned}
	\right.
\end{align*}
As the entries of $y_K$ are strictly bounded between 0 and 1, the error sign is given by
\begin{align*}
	\sign{e_c} =
	\left \lbrace
	\begin{aligned}
		&1 &&\quad \text{if} \quad c=c^*,\\
		-&1		&&\quad \text{otherwise.}
	\end{aligned}
	\right.
\end{align*}

Due to the nonlinearity in the output layer forcing the output values to remain strictly bounded between 0 and 1, the error sign is class-dependent and known in advance as training examples ($x$,$\:c^*$) already provide the error sign information with the label $c^*$. A feedback pathway is thus no longer required as we have shown that the error sign allows providing useful modulatory signals to train multi-layer networks. Therefore, beyond being free from the \textit{weight transport problem} as DFA, sDFA also allows releasing \textit{update locking} and the associated memory overhead in classification problems.

\vspace*{2mm}
\subsection{Direct random target projection delivers useful modulatory signals for classification} \label{ssec_drtp_useful}

This section provides the grounds to show why the proposed direct random target projection (DRTP) algorithm delivers useful modulatory signals to multi-layer networks in the framework of classification problems. First, we show how DRTP can be viewed as a simplified version of sDFA in which the target vector $y^*$ is used as a surrogate for the error sign. Next, we demonstrate mathematically that, in a multi-layer network composed of linear hidden layers and a nonlinear output layer, the modulatory signals prescribed by DRTP and BP are always within 90$^\circ$ of each other, thus providing learning in multi-layer networks.

\begin{algorithm}[!t]
\caption{\textbf{Pseudocode for the direct random target projection (DRTP) algorithm.} $k \in [1, K]$, $k \in \mathbb{N}$, denotes the layer index and $W_k$, $b_k$, $B_k$ and $f_k(\cdot)$ denote the trainable forward weights and biases, the fixed random connectivity matrices and the activation function of the $k$-th hidden layer, respectively. The weighted sum of inputs or pre-activation is denoted as $z_k$ and the layer output or post-activation is denoted as $y_k$, with $y_0$ corresponding to the input $x$. The one-hot-encoding of labels among $C$ output classes is denoted as $y^{*}$ and the learning rate as $\eta$. The update for the weights and biases in the output layer are computed for sigmoid/softmax output units with a binary/categorical cross-entropy loss.}
\label{algo:DRTP}
\begin{algorithmic}
\FOR{$(k = 1; k\le K; k = k+1)$}
\vspace{0.1cm}
\STATE $z_k \leftarrow W_k y_{k-1} + b_k$
\STATE $y_k \leftarrow f_k(z_k)$
\vspace{0.25cm}
\IF{$k < K$}
\vspace{0.1cm}
\STATE $W_k \leftarrow W_k + \eta \left( B_k^T y^{*} \odot f_k^{'}(z_k)\right) y_{k-1}^T$
\STATE $b_k \leftarrow b_k + \eta \left( B_k^T y^{*} \odot f_k^{'}(z_k)\right)$
\vspace{0.1cm}
\ELSE
\vspace{0.1cm}
\STATE $W_K \leftarrow W_K + \frac{\eta}{C} (y^{*}-y_K) y_{k-1}^T$
\STATE $b_K \leftarrow b_K + \frac{\eta}{C} (y^{*}-y_K)$
\vspace{0.1cm}
\ENDIF
\ENDFOR
\end{algorithmic}
\end{algorithm}

\paragraph{DRTP is a simplified version of error-sign-based DFA.} As we have shown that sDFA solves both the \textit{weight transport} and the \textit{update locking} problems in classification tasks, we propose the direct random target projection (DRTP) algorithm, illustrated in Fig.~\ref{fig:topologies}D and written in pseudocode in Algorithm \ref{algo:DRTP}, as a simplified version of sDFA that enhances both performance and computational efficiency. In sDFA, the feedback signal randomly projected to the hidden layers is the sign of the error vector $e = y^* - y_K$, while in DRTP, this feedback signal is replaced by the target vector $y^*$. Being a one-hot encoding of $c^*$, $y^*$ has a single positive entry corresponding to the correct class and zero entries elsewhere:

\begin{align*}
	y^*_c = \frac{1+\text{sign}(e_c)}{2} = 
	\left \lbrace
	\begin{aligned}
		&1      &&\quad \text{if} \quad c=c^*,\\
		&0		&&\quad \text{otherwise.}
	\end{aligned}
	\right.
\end{align*}

Thus, $y^*$ corresponds to a surrogate for the error sign vector used in sDFA, where shift and rescaling operations have been applied to $\text{sign}(e)$. As the connectivity matrices $B_k$ in the DRTP gradients $\delta y_k = B_k^T y^*$ are fixed and random~(Figure~\ref{fig:topologies}D), they can be viewed as comprising the rescaling operation. Only the shift operation applied to $\text{sign}(e)$ makes a critical difference between DRTP and sDFA, which is favorable to DRTP for two reasons. First, DRTP is computationally cheaper than sDFA. Indeed, projecting the target vector $y^*$ to the hidden layers through fixed random connectivity matrices is equivalent to a label-dependent selection of a layerwise random vector. On the contrary, sDFA requires multiplying the error sign vector with the fixed random connectivity matrices for each training example, as all entries of the error sign vector are non-zero. Second, experiments on the MNIST and CIFAR-10 datasets show that DRTP systematically outperforms sDFA (Supplementary Figures~S\ref{fig:mnist_drtp_sdfa}A and S\ref{fig:cifar10_drtp_sdfa}A, Supplementary Tables~S\ref{table:mnist_supplementary} and S\ref{table:cifar10_supplementary}). Indeed, when the feedback information only relies on the error sign and no longer on its magnitude, the weight updates become less selective to the useful information: as all entries of the error sign vector have unit norm, the $C-1$ entries corresponding to incorrect classes outweigh the single entry associated to the correct class and degrade the alignment (Supplementary Figures~S\ref{fig:mnist_drtp_sdfa}B and S\ref{fig:cifar10_drtp_sdfa}B).

\paragraph{The directions of the DRTP and BP modulatory signals are within 90$^\circ$ of each other.} We provide a mathematical proof of alignment between the DRTP and BP modulatory signals. The structure of our proof is inspired from the FA proof of alignment in \cite{Lillicrap16}, which we expand in two ways. First, we extend this proof for the case of DRTP. Second, while \cite{Lillicrap16} demonstrate the alignment with the BP modulatory signals for a network consisting of a single linear hidden layer, a linear output layer and a mean squared error loss, we demonstrate that alignment can be achieved for an arbitrary number of linear hidden layers, a nonlinear output layer with sigmoid/softmax activation and a binary/categorical cross-entropy loss for classification problems. Both proofs are restricted to the case of a single training example. Under these conditions, it is possible to guarantee that the DRTP modulatory signals are aligned with those of BP. This comes from the fact that the prescribed weight updates lead to a soft alignment between the product of forward weight matrices and the fixed random connectivity matrices. The mathematical details, including the lemma and theorem proofs, have been abstracted out to Supplementary Note~\ref{sec_supplementary_proof}.

\begin{figure}[!t]
	\centering
	\includegraphics[width=.3\textwidth]{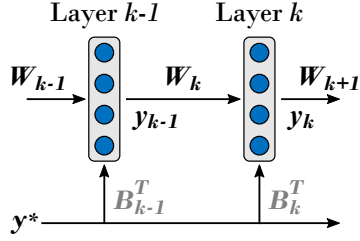}
	\caption{\textbf{Network of DRTP-updated linear hidden layers considered in the context of the mathematical proof of alignment between the DRTP and BP modulatory signals.} The same conventions as in Figure~\ref{fig:topologies} are used.}
	\label{fig:proof_illustration}
\end{figure}

In the case of the multi-layer neural network composed of linear hidden layers shown in Figure~\ref{fig:proof_illustration}, the output of the $k$-th hidden layer is given by
\begin{equation*}
	y_k = z_k = W_k y_{k-1} \quad \text{for} \: k \in [1,K-1],
\end{equation*}
where $K$ is the number of layers, $y_0 = x$ is the input vector, and the bias vector $b_k$ is omitted without loss of generality. The output layer is described by
\begin{align*}	
	\begin{aligned}
		z_K &= W_K y_{K-1},\\
		y_K &= \sigma\left(z_K\right),
	\end{aligned}
\end{align*}
where $\sigma(\cdot)$ is either the sigmoid or the softmax activation function. The loss function $J(\cdot)$ is either the binary cross-entropy (BCE) loss for sigmoid output units or the categorical cross-entropy (CCE) loss for softmax output units, computed over the $C$ output classes:
\begin{align*}
	J_{\text{BCE}}(y_K, y^*) &= -\frac{1}{C} \sum_{c=1}^{C} \Big( y_c^* \log\left(y_{Kc}\right) + (1-y_c^*) \log\left(1-y_{Kc}\right) \Big),\\
	J_{\text{CCE}}(y_K, y^*) &= -\frac{1}{C} \sum_{c=1}^{C} \Big( y_c^* \log\left(y_{Kc}\right) \Big).
\end{align*}

\textit{\textbf{Lemma.~~}} In the case of zero-initialized weights, i.e.~$W_k^0 = 0$ for $k \in [1,K]$, $k \in \mathbb{N}$, and hence of zero-initialized hidden layer outputs, i.e.~$y_k^0=0$ for $k \in [1,K-1]$ and $z_K^0 = 0$, considering a DRTP-based training performed recursively with a single element of the training set $(x,c^*)$ and $y^*$ denoting the one-hot encoding of $c^*$, at every discrete update step $t$, there are non-negative scalars $s_{y_k}^t$ and $s_{W_k}^t$ for $k \in [1,K-1]$ and a $C$-dimensional vector $s_{W_K}^t$ such that

\begin{align*}
	\begin{alignedat}{4}
		&y_k^t &&= -&&s_{y_k}^t \left( B_k^T y^* \right) \qquad &&\text{for} \qquad k \in [1,K-1]\\
		&W_1^t &&= -&&s_{W_1}^t \left( B_1^T y^* \right) x^T &&\\
		&W_k^t &&=  &&s_{W_k}^t \left( B_k^T y^* \right) \left( B_{k-1}^T y^* \right)^T \qquad &&\text{for} \qquad k \in [2,K-1]\\
		&W_K^t &&= -&&s_{W_K}^t \left( B_{K-1}^T y^* \right)^T. &&
	\end{alignedat}
\end{align*}

\textit{\textbf{Theorem.~~}} Under the same conditions as in the lemma and for the linear-hidden-layer network dynamics described above, the $k$-th layer modulatory signals prescribed by DRTP are always a negative scalar multiple of the Moore-Penrose pseudo-inverse of the product of forward matrices of layers $k+1$ to $K$, located in the feedback pathway between the output layer and the $k$-th hidden layer, multiplied by the error. That is, for $k \in [1,K-1]$ and $t>0$,

\begin{equation*}
	- \frac{1}{s_k^t} \left( \prod_{i=K}^{k+1} W_i^t \right)^+ e = B_k^T y^* \quad \text{with} \quad s_k^t > 0.
\end{equation*}

\textit{\textbf{Alignment.~~}} In the framework of classification problems, as the coefficients $s_k^t$ are strictly positive scalars for $t>0$, it results from the theorem that the dot product between the BP and DRTP modulatory signals is strictly positive,~i.e.

\begin{align*}
	\begin{aligned}
	- e^T \left( \prod_{i=k+1}^K W_i^T \right)^T \left( B_k^T y^*\right) &> 0\\
	e^T \underbrace{\left( \prod_{i=k+1}^K W_i^T \right)^T  \left( \prod_{i=K}^{k+1} W_i \right)^+}_{I} \frac{e}{s_k^t} &> 0\\
	\frac{e^Te}{s_k^t} &> 0.
	\end{aligned}
\end{align*}

The BP and DRTP modulatory signals are thus within 90$^\circ$ of each other. \qed

\vspace*{2mm}
\subsection{DRTP learns to classify MNIST and CIFAR-10 images without feedback} \label{ssec:results_mnist_cifar10}

In this section, we compare DRTP with BP and other feedback-alignment-based algorithms, namely FA and DFA, on the MNIST and CIFAR-10 datasets. Both datasets have 10 output classes, they respectively consist in classifying 28$\times$28 grayscale images of handwritten digits for MNIST and 32$\times$32 RGB images of vehicles and animals for CIFAR-10. The network topologies considered in our experiments are, on the one hand, fully-connected (FC) networks with one or two hidden layers, respectively denoted as FC1 and FC2, each hidden layer being constituted of either 500 or 1000 tanh units. On the other hand, convolutional (CONV) networks are used with either fixed random or trainable kernels. The CONV network for MNIST consists of one convolutional layer followed by a max-pooling layer and one fully-connected hidden layer, while for CIFAR-10 it consists of two convolutional layers, each followed by a max-pooling layer, and two fully-connected hidden layers (see Section~\ref{sec_methods}).
 
\subsubsection{MNIST}\label{sssec:results_mnist}

\begin{table}[!ht]
\vspace*{2mm}
\caption{\textbf{Mean and standard deviation of the test error on the MNIST dataset over 10 trials.} DO stands for dropout and indicates the dropout probability used in the fully-connected layers of both FC and CONV networks. The FC networks consist of one (FC1) or two (FC2) hidden layers comprising 500 or 1000 tanh units, with an output fully-connected layer of 10 sigmoid units. The CONV network topology is as follows: a convolutional layer with 32 5$\times$5 kernels, a stride of 1 and a padding of 2, a max-pooling layer with 2$\times$2 kernels and a stride of 2, a fully-connected layer of 1000 tanh units and an output fully-connected layer of 10 sigmoid units.}
\label{table:mnist}
\centering
\begin{tabular}{ccccccc}
\toprule
Network & & BP & FA & DFA & DRTP & Shallow\\
\midrule
\multirow{3}{*}{FC1-500} & DO 0.0 & 1.65$\pm$0.06\% & 1.71$\pm$0.05\% & 1.76$\pm$0.05\% & 4.61$\pm$0.13\% & 8.25$\pm$0.09\% \\
& DO 0.1 & 1.59$\pm$0.03\% & 1.63$\pm$0.05\% & 1.68$\pm$0.03\% & 4.92$\pm$0.13\% & 9.17$\pm$0.11\% \\
& DO 0.25 & 1.76$\pm$0.05\% & 1.74$\pm$0.04\% & 1.86$\pm$0.03\% & 5.75$\pm$0.09\% & 10.15$\pm$0.11\% \\
\midrule
\multirow{3}{*}{FC1-1000} & DO 0.0 & 1.57$\pm$0.04\% & 1.62$\pm$0.05\% & 1.67$\pm$0.03\% & 4.10$\pm$0.07\% & 7.92$\pm$0.10\% \\
& DO 0.1 & 1.48$\pm$0.03\% & 1.55$\pm$0.05\% & 1.58$\pm$0.05\% & 4.31$\pm$0.06\% & 9.29$\pm$0.12\% \\
& DO 0.25 & 1.54$\pm$0.04\% & 1.56$\pm$0.02\% & 1.63$\pm$0.03\% & 4.94$\pm$0.06\% & 10.01$\pm$0.17\% \\
\midrule
\multirow{3}{*}{FC2-500} & DO 0.0 & 1.46$\pm$0.08\% & 1.72$\pm$0.04\% & 1.69$\pm$0.06\% & 4.58$\pm$0.09\% & 8.25$\pm$0.10\% \\
& DO 0.1 & 1.46$\pm$0.04\% & 1.51$\pm$0.04\% & 1.57$\pm$0.06\% & 5.00$\pm$0.07\% & 9.33$\pm$0.09\% \\
& DO 0.25 & 1.38$\pm$0.04\% & 1.69$\pm$0.02\% & 1.52$\pm$0.03\% & 5.94$\pm$0.06\% & 11.01$\pm$0.12\% \\
\midrule
\multirow{3}{*}{FC2-1000} & DO 0.0 & 1.50$\pm$0.09\% & 1.57$\pm$0.06\% & 1.65$\pm$0.07\% & 4.00$\pm$0.10\% & 7.85$\pm$0.09\% \\
& DO 0.1 & 1.46$\pm$0.02\% & 1.46$\pm$0.03\% & 1.57$\pm$0.03\% & 4.25$\pm$0.06\% & 8.73$\pm$0.08\%  \\
& DO 0.25 & 1.38$\pm$0.03\% & 1.50$\pm$0.05\% & 1.45$\pm$0.03\% & 5.05$\pm$0.09\% & 9.84$\pm$0.05\% \\
\midrule
\multirow{3}{*}{\shortstack{CONV\\(random)}} & DO 0.0 & 1.21$\pm$0.05\% & 1.30$\pm$0.06\% & 1.25$\pm$0.08\% & 1.82$\pm$0.11\% & 2.83$\pm$0.19\% \\
& DO 0.1 & 1.25$\pm$0.03\% & 1.33$\pm$0.06\% & 1.30$\pm$0.06\% & 2.06$\pm$0.08\% & 4.74$\pm$0.30\% \\
& DO 0.25 & 1.29$\pm$0.04\% & 1.32$\pm$0.06\% & 1.33$\pm$0.05\% & 2.60$\pm$0.14\% & 6.49$\pm$0.35\% \\
\midrule
\multirow{3}{*}{\shortstack{CONV\\(trained)}} & DO 0.0 & 0.93$\pm$0.04\% & 1.22$\pm$0.06\% & 1.31$\pm$0.06\% & 1.48$\pm$0.15\% &  \\
& DO 0.1 & 1.03$\pm$0.04\% & 1.27$\pm$0.06\% & 1.34$\pm$0.06\% & 1.50$\pm$0.17\% & -- \\
& DO 0.25 & 1.00$\pm$0.03\% & 1.29$\pm$0.04\% & 1.40$\pm$0.06\% & 1.81$\pm$0.20\% &  \\
\bottomrule
\end{tabular}
\vspace*{2mm}
\end{table}

The results on the MNIST dataset are summarized in Table~\ref{table:mnist}. In FC networks, BP, FA and DFA perform similarly, the accuracy degradation of FA and DFA is marginal. While there is a higher accuracy degradation for DRTP, it compares favorably to shallow learning, which suffers from a high accuracy penalty. It shows that DRTP allows training hidden layers to learn MNIST digit classification without feedback. The CONV network topology leads to the lowest error, highlighting that extracting spatial information, even with random kernels, is sufficient to solve the MNIST task. The accuracy slightly degrades along the FA, DFA and DRTP algorithms, with a higher gap for shallow learning. When kernels are trained, BP provides the highest improvement compared to the error obtained with random kernels, followed by DRTP, while no significant change can be observed for FA and DFA. This is likely due to the fact that there is not enough parameter redundancy in convolutional layers to allow for an efficient training with feedback-alignment-based algorithms, which is commonly referred to as a \textit{bottleneck effect} (see Section~\ref{sec_discussion}). Indeed, the angle between the BP loss gradients and the feedback-alignment-based ones is roughly 90$^\circ$, leading to random updates (Supplementary Figure~S\ref{fig:mnist_conv_angle}). This improved performance of DRTP with trained kernels is thus unexpected. Regarding dropout, a positive impact is shown on BP, FA and DFA: a moderate dropout probability is beneficial for FC1 networks, while increasing it to 0.25 can be used for FC2 networks. Dropout has no positive impact for CONV networks, while it degrades the accuracy obtained with DRTP and shallow learning in all cases.

\subsubsection{CIFAR-10}\label{sssec:results_cifar10}

\begin{table}[!ht]
\vspace*{2mm}
\caption{\textbf{Mean and standard deviation of the test error on the CIFAR-10 dataset over 10 trials.} DO stands for dropout and indicates the dropout probability used in the fully-connected layers of both FC and CONV networks. DA stands for data augmentation, which consists in horizontally flipping the training images. No dropout is used for DA. The FC networks consist of one (FC1) or two (FC2) hidden layers comprising 500 or 1000 tanh units, with an output fully-connected layer of 10 sigmoid units. The CONV network topology is as follows: two convolutional layers with respectively 64 and 256 3$\times$3 kernels, a stride and a padding of 1, both followed by a max-pooling layer with 2$\times$2 kernels and a stride of 2, then two fully-connected layers of 1000 tanh units and an output fully-connected layer of 10 sigmoid units.}
\label{table:cifar10}
\centering
\begin{tabular}{ccccccc}
\toprule
Network & & BP & FA & DFA & DRTP & Shallow \\
\midrule
\multirow{4}{*}{FC1-500} & DO 0.0 & 48.45$\pm$0.38\% & 49.38$\pm$0.22\% & 49.62$\pm$0.29\% & 53.92$\pm$0.23\% & 58.83$\pm$0.27\% \\
& DO 0.1 & 47.48$\pm$0.39\% & 48.94$\pm$0.22\% & 48.85$\pm$0.23\% & 53.77$\pm$0.17\% & 59.33$\pm$0.17\% \\
& DO 0.25 & 47.80$\pm$0.21\% & 48.62$\pm$0.23\% & 48.65$\pm$0.29\% & 54.26$\pm$0.16\% & 60.44$\pm$0.14\% \\
& DA & 45.87$\pm$0.22\% & 47.11$\pm$0.34\% & 47.34$\pm$0.26\% & 52.73$\pm$0.31\% & 58.60$\pm$0.20\% \\
\midrule
\multirow{4}{*}{FC1-1000} & DO 0.0 & 47.52$\pm$0.30\% & 48.47$\pm$0.18\% & 48.44$\pm$0.34\% & 53.34$\pm$0.10\% & 57.91$\pm$0.17\% \\
& DO 0.1 & 46.42$\pm$0.28\% & 47.72$\pm$0.19\% & 47.79$\pm$0.31\% & 53.15$\pm$0.15\% & 58.35$\pm$0.24\% \\
& DO 0.25 & 46.21$\pm$0.16\% & 47.11$\pm$0.18\% & 47.11$\pm$0.25\% & 53.39$\pm$0.15\% & 59.20$\pm$0.18\% \\
& DA & 45.01$\pm$0.33\% & 46.15$\pm$0.36\% & 46.24$\pm$0.32\% & 51.87$\pm$0.32\% & 57.40$\pm$0.24\% \\
\midrule
\multirow{4}{*}{FC2-500} & DO 0.0 & 49.03$\pm$0.22\% & 50.66$\pm$0.24\% & 50.45$\pm$0.36\% & 53.41$\pm$0.35\% & 59.62$\pm$0.34\% \\
& DO 0.1 & 48.32$\pm$0.16\% & 49.64$\pm$0.23\% & 49.58$\pm$0.30\% & 54.06$\pm$0.46\% & 60.34$\pm$0.24\% \\
& DO 0.25 & 49.96$\pm$0.18\% & 50.80$\pm$0.16\% & 50.00$\pm$0.13\% & 54.57$\pm$0.33\% & 61.77$\pm$0.18\% \\
& DA & 46.62$\pm$0.10\% & 48.55$\pm$0.25\% & 48.75$\pm$0.28\% & 52.54$\pm$0.34\% & 58.99$\pm$0.20\% \\
\midrule
\multirow{4}{*}{FC2-1000} & DO 0.0 & 48.81$\pm$0.22\% & 49.87$\pm$0.18\% & 50.05$\pm$0.21\% & 52.68$\pm$0.25\% & 58.59$\pm$0.14\% \\
& DO 0.1 & 46.58$\pm$0.24\% & 47.97$\pm$0.18\% & 48.68$\pm$0.34\% & 52.45$\pm$0.15\% & 59.12$\pm$0.13\% \\
& DO 0.25 & 47.65$\pm$0.16\% & 48.82$\pm$0.15\% & 48.08$\pm$0.14\% & 53.29$\pm$0.31\% & 60.52$\pm$0.14\% \\
& DA & 46.05$\pm$0.14\% & 47.41$\pm$0.27\% & 47.90$\pm$0.19\% & 51.27$\pm$0.21\% & 57.90$\pm$0.23\% \\
\midrule
\multirow{4}{*}{\shortstack{CONV\\(random)}} & DO 0.0 & 29.83$\pm$0.25\% & 30.27$\pm$0.45\% & 29.98$\pm$0.30\% & 32.65$\pm$0.38\% & 44.89$\pm$0.67\% \\
& DO 0.1 & 29.49$\pm$0.36\% & 29.58$\pm$0.33\% & 29.44$\pm$0.31\% & 32.57$\pm$0.34\% & 48.38$\pm$0.33\% \\
& DO 0.25 & 30.39$\pm$0.32\% & 30.55$\pm$0.28\% & 30.31$\pm$0.35\% & 33.90$\pm$0.53\% & 52.27$\pm$0.34\% \\
& DA & 27.87$\pm$0.25\% & 28.52$\pm$0.40\% & 28.46$\pm$0.43\% & 31.04$\pm$0.45\% & 44.23$\pm$0.42\% \\
\midrule
\multirow{4}{*}{\shortstack{CONV\\(trained)}} & DO 0.0 & 25.31$\pm$0.25\% & 29.92$\pm$0.26\% & 31.38$\pm$0.38\% & 35.82$\pm$0.59\% & \multirow{4}{*}{--} \\
& DO 0.1 & 27.12$\pm$0.23\% & 28.98$\pm$0.36\% & 30.56$\pm$0.41\% & 35.17$\pm$0.91\% & \\
& DO 0.25 & 25.61$\pm$0.23\% & 28.95$\pm$0.17\% & 31.23$\pm$0.38\% & 35.51$\pm$0.61\% & \\
& DA & 25.27$\pm$0.26\% & 28.16$\pm$0.45\% & 29.49$\pm$0.49\% & 34.39$\pm$0.64\% & \\
\bottomrule
\end{tabular}
\vspace*{2mm}
\end{table}

The results on the CIFAR-10 dataset are summarized in Table \ref{table:cifar10}, highlighting conclusions similar to those already drawn for the MNIST dataset. Compared to BP, accuracy degrades along the FA, DFA and DRTP algorithms. The gap is higher for DRTP, yet it again compares favorably to shallow learning, demonstrating that DRTP also allows training hidden layers to learn CIFAR-10 image classification without feedback. For CONV networks, if kernels are trained, only BP is able to provide a significant advantage. Due to the bottleneck effect, FA only provides a slight improvement, while DFA and DRTP are negatively impacted. Regarding dropout, a moderate probability of 0.1 works fairly well for BP, FA, DFA and DRTP, while a higher probability of 0.25 rarely provides any advantage. Dropout always leads to an accuracy reduction for shallow learning. Finally, data augmentation~(DA) improves the accuracy of all algorithms and is more effective than dropout.

\vspace*{2mm}
\section{Discussion}\label{sec_discussion}

While the backpropagation of error algorithm allowed taking artificial neural networks to outperform humans on complex datasets such as ImageNet~\cite{He15}, the key problems of \textit{weight transport} and \textit{update locking} highlight how aiming at breaking accuracy records on standard datasets has diverted attention from hardware efficiency considerations. While accuracy is the key driver for applications that can be backed by significant GPU and CPU resources, the development of decentralized adaptive smart sensors calls for keeping hardware requirements of learning algorithms to a minimum. Moreover, it has been shown that weight transport and update locking are not biologically plausible~\cite{Grossberg87,Baldi18}, following from the non-locality in both weight and gradient information. Therefore, there is currently an increasing interest in releasing these constraints in order to achieve higher hardware efficiency and to understand the mechanisms that could underlie biological synaptic plasticity.

The proposed DRTP algorithm successfully addresses both the weight transport and the update locking problems, which has only been partially demonstrated in previously-proposed approaches. Indeed, the FA and DFA algorithms only address the weight transport problem~\cite{Lillicrap16,Nokland16}. The error locality approach still suffers from the weight transport problem in the local classifiers~\cite{Mostafa18,Kaiser18,Nokland19}, while the synthetic gradients approach requires backpropagating gradient information from deeper layers in order to train the layerwise gradient predictors~\cite{Jaderberg17,Czarnecki17}. Both the error locality and the synthetic gradients approaches also incur computational overhead by requiring the addition of side local networks for error or gradient prediction. On the contrary, DRTP is a strikingly simple rule that alleviates the two key BP issues by enabling each layer to be updated with local information as the forward evaluation proceeds. In order to estimate the layerwise loss gradients $\delta y_k$ for each layer, the only operation required by DRTP is a label-dependent random vector selection~(Figure~\ref{fig:topologies}\textbf{D}). Despite the absence of dedicated feedback pathways, we demonstrated on the MNIST and CIFAR-10 datasets that DRTP allows training hidden layers at low computational and memory costs, thus highlighting its suitability for deployment in adaptive smart sensors at the edge and for embedded systems in general. In terms of floating-point operations (FLOPs), the overhead of DRTP weight updates is approximately equal to the cost of the forward pass, assuming that (i) the number of classes of the problem is negligible compared to the number of units in the hidden layers, which is typical of edge computing tasks, and (ii) the learning rate is embedded in the magnitude of the random connectivity matrices $B_k^T$. Doubling the computational cost of shallow-learning networks (i.e. doubling the numbers of hidden units or hidden layers) does not allow recovering their performance gap compared to DRTP-updated networks (Tables \ref{table:mnist} and \ref{table:cifar10}). Even more importantly when considering dedicated hardware implementations for edge computing, the memory requirements should be minimized so as to fit the whole network topology into on-chip memory resources. Indeed, accesses to off-chip DRAM memory are three orders of magnitude more expensive energy-wise than a 32-bit FLOP~\cite{Horowitz14}. Therefore, as opposed to increasing the resources of shallow-trained networks, DRTP offers a low-overhead training algorithm operating on small network topologies, ideally suiting edge-computing hardware requirements. These claims are proven \textit{in silico} in \cite{Frenkel20}, where implementing DRTP in an event-driven convolutional processor requires only 16.8-\% power and 11.8-\% silicon area overheads and allows demonstrating a favorable accuracy-power-area tradeoff compared to both on-chip online- and off-chip offline-trained conventional machine learning accelerators on the MNIST dataset.

By solving the weight transport and update locking problems, DRTP also releases key biological implausibility issues. Neurons in the brain separate forward and backward information in somatic and dendritic compartments, a property that is highlighted in the formulation of three-factor synaptic plasticity rules~\cite{Urbanczik14}: pre-synaptic and post-synaptic activities are modulated by a third factor corresponding to a local dendritic voltage. \cite{Lillicrap16} build on the idea that a separate dendritic compartment integrates higher-order feedback and generates local teaching signals, where the errors could be viewed as a mismatch between expected and actual perceptions or actions. This aspect is further emphasized in the subsequent work of~\cite{Guerguiev17} when framing DFA as a spike-based three-factor learning rule. In the case of DRTP, compared to DFA, the error signal is replaced by the targets, which could correspond to a modulation that bypasses the actual perceptions or realized actions, relying only on predictions or intentions. Furthermore, DRTP could come in line with recent findings in cortical areas that reveal the existence of output-independent target signals in the dendritic instructive pathways of intermediate-layer neurons~\cite{Magee20}. Understanding the mechanisms of synaptic plasticity is critical in the field of neuromorphic engineering, which aims at porting biological computational principles to hardware toward higher energy efficiency~\cite{Thakur18,Rajendran19}. However, even simple local bio-inspired learning rules such as spike-timing-dependent plasticity (STDP)~\cite{Bi98} can lead to non-trivial hardware requirements, which currently hinders adaptive neuromorphic systems from reaching high-density large-scale integration~\cite{Frenkel19a}. While adaptations of STDP, such as spike-dependent synaptic plasticity (SDSP)~\cite{Brader07}, release most of the STDP hardware constraints, their training performance is currently not sufficient to support deployability of neuromorphic hardware for real-world scenarios~\cite{Frenkel19a,Frenkel19b}. A three-factor formulation of DRTP would release the update locking problem in the spike-based three-factor formulations of DFA~\cite{Guerguiev17,Neftci17}, which currently imply memory and control overhead in their hardware implementations~\cite{Detorakis18,Park19}. Porting DRTP to neuromorphic hardware is thus a natural next step.

While DRTP relaxes structural, memory and computational requirements toward decentralized hardware deployment, the accuracy degradation over DFA comes from the fact that only the error sign is taken into account, not its class-dependent magnitude. This could be mitigated by keeping track of the error magnitude over the last samples in order to modulate the layerwise learning rates, at the expense of releasing the purely feedforward nature of DRTP. A learning rate scheduler could also be used. The DRTP algorithm was derived specifically for classification problems with sigmoid/softmax output units and a binary/categorical cross-entropy loss, yet hidden layer activations also play a key role in the learning dynamics of DRTP. As the estimated loss gradients $\delta y_{k}$ computed from the targets have a constant sign and magnitude, the weights updates only change due to the previous layer outputs and the derivative of the activation function, as training progresses. When using activation functions such as tanh in the hidden layers, the network stops learning thanks to the activation function derivative, whose value vanishes as its input argument moves away from zero. This mechanism specific to DRTP is highlighted in Supplementary Figures~S\ref{fig:mnist_activations}-S\ref{fig:final_activations} and could be exploited to generate networks whose activations can be binarized during inference, which we will investigate in future work. In return, only activation functions presenting this saturation property are expected to lead to satisfying performance when used in conjunction with DRTP, which for example excludes ReLU activations.

Finally, as for all other feedback-alignment-based algorithms, DRTP only slightly improves or even degrades the accuracy when applied to convolutional layers. Convolutional layers do not provide the parameter redundancy that can be found in fully-connected layers, a \textit{bottleneck effect} that was first highlighted for FA~\cite{Lillicrap16} and has recently been studied for DFA~\cite{Launay19}. Nevertheless, other training algorithms based either on a greedy layerwise learning \cite{Belilovsky19} or on the alignment with local targets \cite{Ororbia19} have proven to be successful in training convolutional layers at the expense of only partially solving the update locking problem. Indeed, the training algorithm proposed in \cite{Belilovsky19} still suffers from update locking in the layerwise auxiliary networks while the one proposed in \cite{Ororbia19} relies on the backpropagation of the output error to compute the layerwise targets. If fixed random convolutional layers do not meet the performance requirements of the target application, a combination of DRTP for fully-connected layers together with error locality or synthetic gradients approaches for convolutional layers can be considered. This granularity in the selection of learning mechanisms, trading off accuracy and hardware efficiency, comes in accordance with the wide spectrum of plasticity mechanisms that are believed to operate in the brain~\cite{Zenke15}.

\vspace*{2mm}
\section{Materials and Methods}\label{sec_methods}

The training on both the synthetic regression and classification tasks and the MNIST and CIFAR-10 datasets has been carried out with PyTorch~\cite{Paszke17}, one of the numerous Python frameworks supporting deep learning. In all experiments, the reported update angles between feedback-alignment-based algorithms and BP were generated at each update step, where the BP update values were computed solely to assess the evolution of the alignment angle over the update steps carried out by FA, DFA, sDFA or DRTP.

\paragraph{Regression.} The examples in the training and test sets are denoted as $(x,y^*)$. The 10-dimensional target vectors $y^*$ are generated using $y^*_j = T_j(x) = \text{cos}(\overline{x}+\phi_j)$, where $\phi_j = -\pi/2 + j\pi/9$ for $j \in [0,9]$ and $j \in \mathbb{N}_0$. $\overline{x}$ denotes the mean of $x$, a 256-dimensional vector whose entries are initialized from a normal distribution with a mean sampled from a uniform distribution between $-\pi$ and $\pi$ and with a unit variance. The training and test sets respectively contain 5k and 1k examples. The trained network has a 256-100-100-10 topology with tanh hidden and output units, whose forward weights are drawn from a He uniform distribution~\cite{He15} and are zero-initialized for feedback-alignment-based algorithms. The random connectivity matrices of feedback-alignment-based algorithms are also drawn from He uniform distributions. The weights are updated after each minibatch of 50 examples, and the network is trained for 500 epochs with a fixed learning rate $\eta = 5 \times 10^{-4}$ for all training algorithms. The loss function is the mean squared error. The losses on the training and test sets and the alignment angles with BP updates are monitored every 1k samples. The experiment is repeated 10 times for each training algorithm, with different network initializations for each experiment run.

\paragraph{Synthetic data classification.} The examples in the training and test sets are generated using the \mbox{\texttt{make\_classification}} function from the Python library \texttt{sklearn}~\cite{Pedregosa11}. The main inputs required by this function are the number of samples to be generated, the number of features $n$ in the input vectors $x$, the number of informative features $n_{inf}$ among the input vectors, the number of classes, the number of clusters per class and a factor \texttt{class\_sep} which conditions the class separation. In this work, we have used \mbox{$n$ = 256} and \mbox{$n_{inf}$ = 128}, ten classes, five clusters per class and $\texttt{class\_sep}=4.5$. Using this set of parameters, the \texttt{make\_classification} function then generates examples by creating for each class clusters of points normally distributed about the vertices of an $n_{inf}$-dimensional hypercube. The remaining features are filled with normally-distributed random noise. The generated examples are then separated into training and test sets of 25k and 5k examples, respectively. The trained network has a 256-500-500-10 topology with tanh hidden units and sigmoid output units. The forward and backward weights initialization, as well as the forward weight updates, are performed as for regression. As this is a classification task, the loss function is the binary cross-entropy loss. The network is trained for 500 epochs with a fixed learning rate $\eta = 5\times10^{-4}$. The losses on the training and test sets and the alignment angles with BP updates are monitored every 2.5k samples. For each training algorithm, the experiment is repeated 10 times with different network initializations.

\paragraph{MNIST and CIFAR-10 images classification.} A fixed learning rate is selected based on a grid search for each training algorithm, dataset and network type (Table~\ref{table:learning_rates}). For both the MNIST and CIFAR-10 experiments, the chosen optimizer is Adam with default parameters. A sigmoid output layer and a binary cross-entropy loss are used for all training algorithms. The entries of the forward weight matrices $W_k$ are initialized with a He uniform distribution, as well as the entries of the fixed random connectivity matrices $B_k$ of feedback-alignment-based algorithms. When used, dropout is applied with the same probability to all fully-connected layers. For MNIST, the networks are trained for 100 epochs with a minibatch size of 60. The CONV network topology consists of a convolutional layer with 32 5$\times$5 kernels, a stride of 1 and a padding of 2, a max-pooling layer with 2$\times$2 kernels and a stride of 2, a fully-connected layer of 1000 tanh units and an output fully-connected layer of 10 units. For CIFAR-10, a minibatch size of 100 is used and early stopping is applied, with a maximum of 200 epochs. The CONV network topology consists of two convolutional layers with respectively 64 and 256 3$\times$3 kernels, a stride and a padding of 1, both followed by a max-pooling layer with 2$\times$2 kernels and a stride of 2, then two fully-connected layers of 1000 tanh units and an output fully-connected layer of 10 units. For all experiments, the test error is averaged over the last 10 epochs of training. The results reported in Tables~\ref{table:mnist},~\ref{table:cifar10},~S\ref{table:mnist_supplementary} and~S\ref{table:cifar10_supplementary} are the mean and standard deviation over 10 trials.

\begin{table}[!ht]
\caption{\textbf{The learning rate values for the MNIST and CIFAR-10 datasets are selected based on a grid search.} A different learning rate is selected for each training algorithm, dataset and network type.}
\label{table:learning_rates}
\centering
\begin{tabular}{cccccccc}
\toprule
Dataset & Network & BP & FA & DFA & sDFA & DRTP & Shallow\\
\midrule
\multirow{4}{*}{MNIST} & FC1 & 1.5$\times$10$^{-4}$ & 5$\times$10$^{-4}$ & 1.5$\times$10$^{-4}$ & 5$\times$10$^{-4}$ & 1.5$\times$10$^{-4}$ & 1.5$\times$10$^{-2}$\\
& FC2 & 5$\times$10$^{-4}$ & 1.5$\times$10$^{-4}$ & 5$\times$10$^{-4}$ & 5$\times$10$^{-4}$ & 1.5$\times$10$^{-4}$ & 5$\times$10$^{-3}$\\
& CONV (rand.) & $5\times$10$^{-5}$ & $1.5\times$10$^{-4}$ & $5\times$10$^{-5}$ & $5\times$10$^{-4}$ & $5\times$10$^{-4}$ & $5\times$10$^{-3}$\\
& CONV (train.) & 5$\times$10$^{-4}$ & 5$\times$10$^{-5}$ & 5$\times$10$^{-5}$ & 1.5$\times$10$^{-4}$ & 1.5$\times$10$^{-4}$ & -- \\
\midrule
\multirow{4}{*}{CIFAR-10} & FC1 & 1.5$\times$10$^{-5}$ & 1.5$\times$10$^{-5}$ & 1.5$\times$10$^{-5}$ & 5$\times$10$^{-5}$ & 1.5$\times$10$^{-4}$ & 1.5$\times$10$^{-4}$\\
& FC2 & 5$\times$10$^{-6}$ & 5$\times$10$^{-6}$ & 5$\times$10$^{-6}$ & 5$\times$10$^{-5}$ & 5$\times$10$^{-5}$ & 5$\times$10$^{-4}$\\
& CONV (rand.) & 5$\times$10$^{-6}$ & 5$\times$10$^{-6}$ & 5$\times$10$^{-6}$ & 1.5$\times$10$^{-4}$ & 1.5$\times$10$^{-4}$ & 1.5$\times$10$^{-3}$\\
& CONV (train.) & $1.5\times$10$^{-4}$ & $5\times$10$^{-6}$ & $5\times$10$^{-6}$ & $1.5\times$10$^{-5}$ & $5\times$10$^{-5}$ & -- \\
\bottomrule
\end{tabular}
\end{table}

\section*{Code Availability}

The PyTorch code allowing to reproduce all results in this study is available open source under the Apache 2.0 license at \url{https://github.com/ChFrenkel/DirectRandomTargetProjection}.

\section*{Data Availability}

The datasets used in this study are publicly available.

\section*{Acknowledgments}

The authors would like to thank Emre Neftci, Giacomo Indiveri, Marian Verhelst, Simon Carbonnelle and Vincent Schellekens for fruitful discussions and Christophe De Vleeschouwer for granting access to a deep learning workstation.

\section*{Funding}

CF was with Universit\'e catholique de Louvain as a Research Fellow from the National Foundation for Scientific Research (FNRS) of Belgium.

\section*{Conflict of Interest Statement}

The authors declare that the research was conducted in the absence of any commercial or financial relationships that could be construed as a potential conflict of interest.

\section*{Author Contributions}

CF developed the main idea. CF and ML derived the mathematical proofs and worked on the simulation experiments. CF, ML and DB wrote the paper. CF and ML contributed equally to this work.

\newpage

\renewcommand{\thesubsection}{Supplementary Note \arabic{subsection}}
\renewcommand{\figurename}{Supplementary Figure}
\renewcommand{\tablename}{Supplementary Table}
\setcounter{figure}{0}    
\setcounter{table}{0}    

\vspace*{\fill}

\begin{figure}[!ht]
	\centering
	\includegraphics[width=\textwidth]{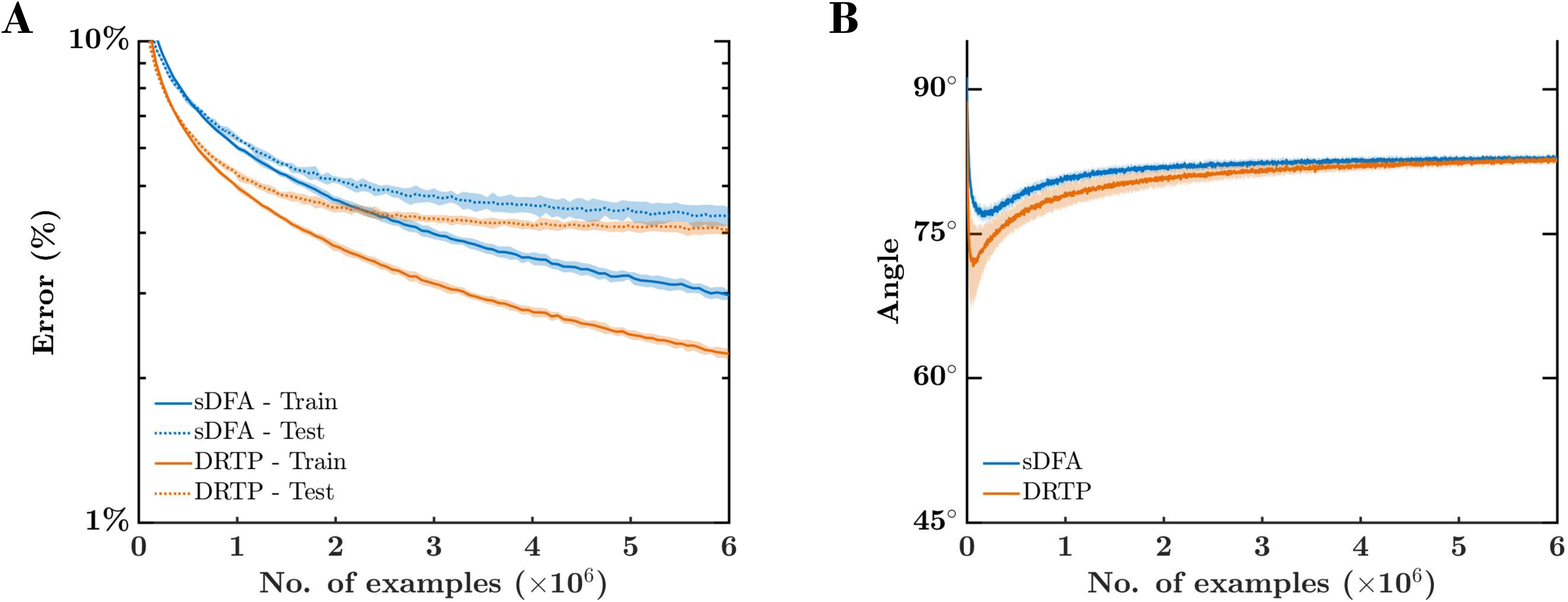}
	\caption{\textbf{DRTP outperforms sDFA on the MNIST dataset.} Both figures are with error bars of one standard deviation over 10 runs. The training and test errors are measured after each epoch, while the angle is measured after each minibatch of 60 examples. Both training methods use Adam with a fixed learning rate of 1.5$\times$10$^{-4}$. \textbf{(A)} A 784-1000-10 network with tanh hidden units and sigmoid output units is trained to classify MNIST handwritten digits with the sDFA and DRTP algorithms. On average, the error on the training set reaches 2.97\% for sDFA and 2.24\% for DRTP, while the error on the test set reaches 4.33\% for sDFA and 4.05\% for DRTP after 100 epochs. \textbf{(B)} While the loss gradients $\delta y_k$ estimated by both sDFA and DRTP are within 90$^\circ$ of the ones prescribed by BP, the alignment angle is initially better for DRTP than for sDFA. The gap vanishes as the training progresses.}
	\label{fig:mnist_drtp_sdfa}
\end{figure}

\vspace*{\fill}

\newpage

\vspace*{\fill}

\begin{figure}[!ht]
	\centering
	\includegraphics[width=\textwidth]{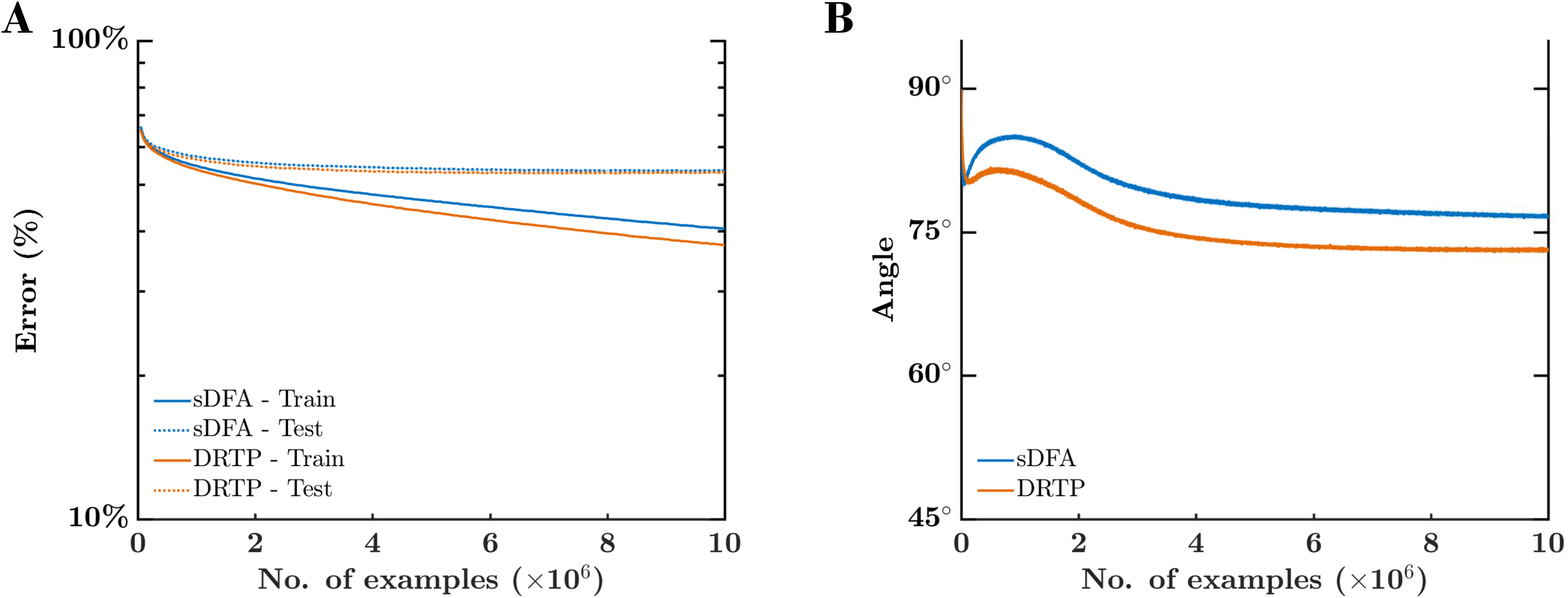}
	\caption{\textbf{DRTP outperforms sDFA on the CIFAR-10 dataset.} Both figures are with error bars of one standard deviation over 10 runs. The training and test errors are measured after each epoch, while the angle is measured after each minibatch of 100 examples. Both training methods use Adam with a fixed learning rate of 5$\times$10$^{-5}$. \textbf{(A)} A 3072-1000-10 network with tanh hidden units and sigmoid output units is trained to classify CIFAR-10 images with the sDFA and DRTP algorithms. On average, the error on the training set reaches 40.74\% for sDFA and 37.39\% for DRTP, while the error on the test set reaches 53.53\% for sDFA and 53.12\% for DRTP after 200 epochs. No early stopping was applied. \textbf{(B)} While the loss gradients $\delta y_k$ estimated by both sDFA and DRTP are within 90$^\circ$ of the ones prescribed by BP, the alignment angle is approximately 3.40$^\circ$ better for DRTP than for sDFA.}
	\label{fig:cifar10_drtp_sdfa}
\end{figure}

\vspace*{\fill}

\newpage

\vspace*{\fill}

\begin{figure}[!ht]
	\centering
	\includegraphics[width=.5\textwidth]{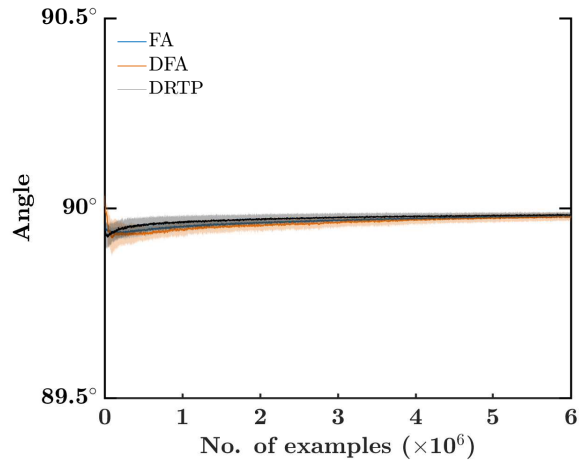}
	\caption{\textbf{Updates to the convolutional layer weights prescribed by feedback-alignment-based algorithms are random due to a 90$^\circ$-alignment with the BP loss gradients $\delta y_k$.} A convolutional network is trained on the MNIST dataset with FA, DFA and DRTP. The network topology and training parameters are identical to those used for the trained CONV network. Error bars are one standard deviation over 10 runs, the angle is measured after each minibatch of 60 examples. Angles have been smoothed by an exponentially-weighted moving average filter with a momentum coefficient of 0.95.}
	\label{fig:mnist_conv_angle}
\end{figure}

\vspace*{\fill}

\newpage

\vspace*{\fill}

\begin{figure}[!ht]
	\centering
	\includegraphics[width=\textwidth]{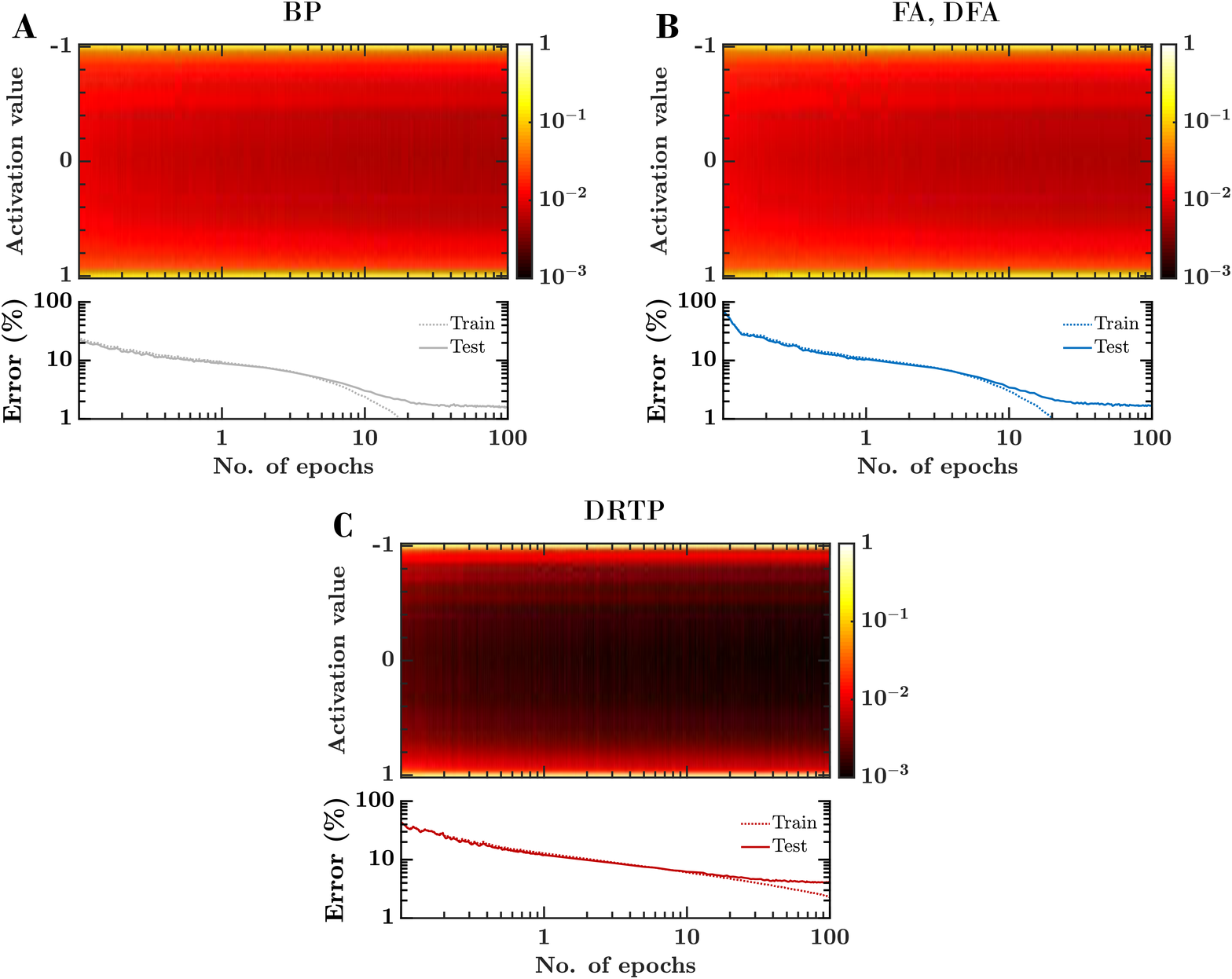}
	\caption{\textbf{On the MNIST dataset, DRTP leads to a distribution of the activation values in the hidden layer that is more heavily skewed towards $\pm$1 than BP, FA and DFA.} A 784-1000-10 network with tanh hidden units and sigmoid output units is trained to classify MNIST images with the BP (\textbf{A}), FA/DFA (\textbf{B}) and DRTP (\textbf{C}) algorithms, where the FA and DFA algorithms are equivalent for single-hidden-layer networks. The network training relies on the Adam optimizer with a binary cross-entropy loss and a fixed learning rate of 1.5$\times$10$^{-4}$, the training and test errors are measured after each minibatch during the first epoch and then after each epoch during the rest of the training. To estimate the probability density function of the activations, their values are monitored for 100 different examples in 100 successive minibatches over the course of training. The estimated probability density function hints at a different learning mechanism for DRTP: as the distribution of the activation values in the hidden layer is more heavily skewed towards $\pm$1, the vanishing value of the tanh activation function derivative in this region leads the network to stop learning.}
	\label{fig:mnist_activations}
\end{figure}

\vspace*{\fill}

\newpage

\vspace*{\fill}

\begin{figure}[!ht]
	\centering
	\includegraphics[width=\textwidth]{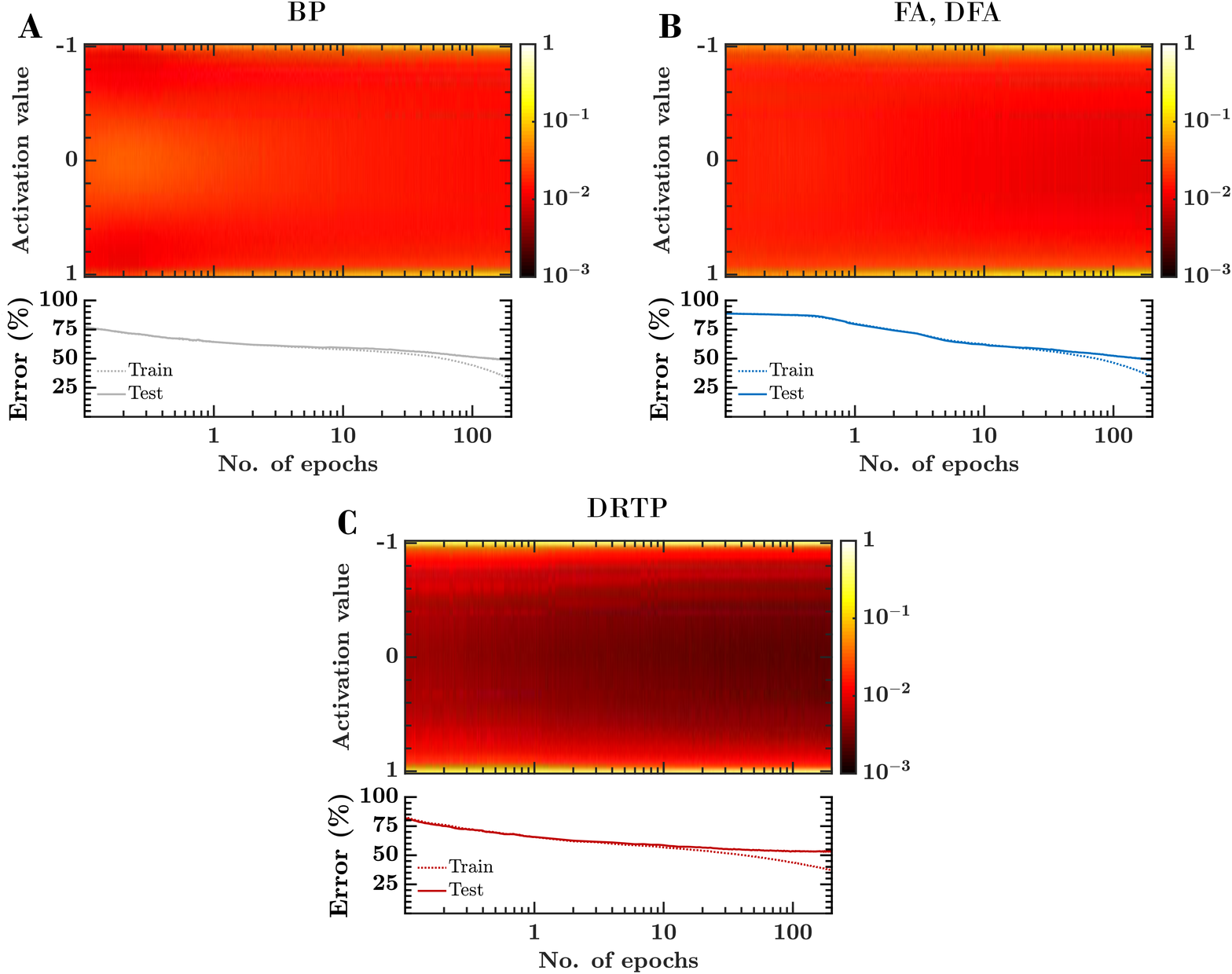}
	\caption{\textbf{On the CIFAR-10 dataset, DRTP leads to a distribution of the activation values in the hidden layer that is more heavily skewed towards $\pm$1 than BP, FA and DFA.} A 3072-1000-10 network with tanh hidden units and sigmoid output units is trained to classify CIFAR-10 images with the BP (\textbf{A}), FA/DFA (\textbf{B}) and DRTP (\textbf{C}) algorithms. The network training relies on the Adam optimizer with a binary cross-entropy loss and a fixed learning rate of 5$\times$10$^{-6}$ for BP and FA/DFA, and 5$\times$10$^{-5}$ for DRTP, as per Table~3 in the main text. Other experimental conditions are similar to the ones used in Fig. \ref{fig:mnist_activations}. Similarly to the experiments on the MNIST dataset, the distribution of the activation values in the hidden layer is more heavily skewed towards $\pm$1 for DRTP, which hints at a stop learning mechanism through the vanishing value of the tanh activation function derivative in this region.}
	\label{fig:cifar10_activations}
\end{figure}

\vspace*{\fill}

\newpage

\vspace*{\fill}

\begin{figure}[!ht]
	\centering
	\includegraphics[width=\textwidth]{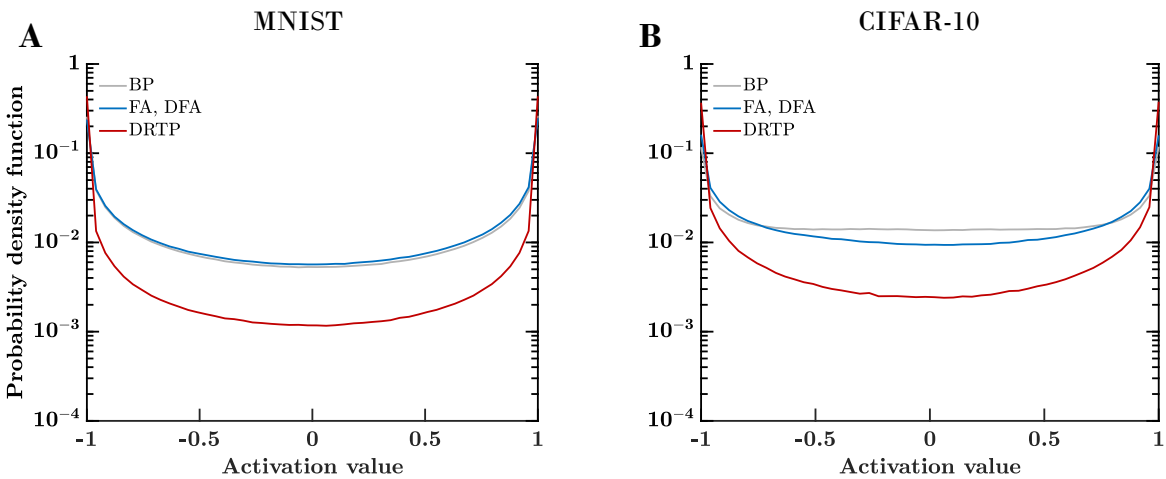}
	\caption{\textbf{The estimated probability density function of the activation values at the end of training exhibits a more pronounced skewing towards $\pm$1 for DRTP than for BP, FA and DFA.} Single-hidden-layer fully-connected networks are trained to classify images from the MNIST (\textbf{A}) and CIFAR-10 (\textbf{B}) datasets, with the experimental conditions described in Figs. \ref{fig:mnist_activations} and \ref{fig:cifar10_activations}.}
	\label{fig:final_activations}
\end{figure}

\vspace*{\fill}

\newpage

\vspace*{\fill}

\begin{table}[!ht]
\caption{\textbf{Comparison of the sDFA and DRTP training algorithms on the MNIST dataset, illustrating that DRTP systematically outperforms sDFA.} The mean and the standard deviation of the test error over 10 trials are provided. The network definitions and conditions are identical to those of Table~\textbf{1}. The learning rates are summarized in Table~\textbf{3}.}
\label{table:mnist_supplementary}
\centering
\begin{tabular}{cccc}
\toprule
Network & & sDFA & DRTP\\
\midrule
\multirow{3}{*}{FC1-500} & DO 0.0 & 4.74$\pm$0.15\% & 4.61$\pm$0.13\% \\
& DO 0.1 & 5.10$\pm$0.13\% & 4.92$\pm$0.13\% \\
& DO 0.25 & 6.06$\pm$0.10\% & 5.75$\pm$0.09\% \\
\midrule
\multirow{3}{*}{FC1-1000} & DO 0.0 & 4.22$\pm$0.11\% & 4.10$\pm$0.07\% \\
& DO 0.1 & 4.42$\pm$0.12\% & 4.31$\pm$0.06\% \\
& DO 0.25 & 5.23$\pm$0.12\% & 4.94$\pm$0.06\% \\
\midrule
\multirow{3}{*}{FC2-500} & DO 0.0 & 4.78$\pm$0.11\% & 4.58$\pm$0.09\% \\
& DO 0.1 & 5.16$\pm$0.13\% & 5.00$\pm$0.07\% \\
& DO 0.25 & 6.13$\pm$0.10\% & 5.94$\pm$0.06\% \\
\midrule
\multirow{3}{*}{FC2-1000} & DO 0.0 & 4.24$\pm$0.09\% & 4.00$\pm$0.10\% \\
& DO 0.1 & 4.51$\pm$0.12\% & 4.25$\pm$0.06\% \\
& DO 0.25 & 5.39$\pm$0.05\% & 5.05$\pm$0.09\% \\
\midrule
\multirow{3}{*}{\shortstack{CONV\\(random)}} & DO 0.0 & 1.88$\pm$0.10\% & 1.82$\pm$0.11\% \\
& DO 0.1 & 2.17$\pm$0.13\% & 2.06$\pm$0.08\% \\
& DO 0.25 & 2.80$\pm$0.17\% & 2.60$\pm$0.14\% \\
\midrule
\multirow{3}{*}{\shortstack{CONV\\(trained)}} & DO 0.0 & 1.69$\pm$0.10\% & 1.48$\pm$0.15\% \\
& DO 0.1 & 1.83$\pm$0.11\% & 1.50$\pm$0.17\% \\
& DO 0.25 & 2.20$\pm$0.15\% & 1.81$\pm$0.20\% \\
\bottomrule
\end{tabular}
\end{table}

\vspace*{\fill}

\newpage

\vspace*{\fill}

\begin{table}[!ht]
\caption{\textbf{Comparison of the sDFA and DRTP training algorithms on the CIFAR-10 dataset, illustrating that DRTP systematically outperforms sDFA.} The mean and the standard deviation of the test error over 10 trials are provided. The network definitions and conditions are identical to those of Table~\textbf{2}. The learning rates are summarized in Table~\textbf{3}.}
\label{table:cifar10_supplementary}
\centering
\begin{tabular}{cccc}
\toprule
Network & & sDFA & DRTP \\
\midrule
\multirow{4}{*}{FC1-500} & DO 0.0 & 54.80$\pm$0.29\% & 53.92$\pm$0.23\% \\
& DO 0.1 & 54.79$\pm$0.24\% & 53.77$\pm$0.17\% \\
& DO 0.25 & 55.48$\pm$0.27\% & 54.26$\pm$0.16\% \\
& DA & 53.83$\pm$0.32\% & 52.73$\pm$0.31\% \\
\midrule
\multirow{4}{*}{FC1-1000} & DO 0.0 & 53.73$\pm$0.33\% & 53.34$\pm$0.10\% \\
& DO 0.1 & 53.92$\pm$0.31\% & 53.15$\pm$0.15\% \\
& DO 0.25 & 54.60$\pm$0.38\% & 53.39$\pm$0.15\% \\
& DA & 52.95$\pm$0.32\% & 51.87$\pm$0.32\% \\
\midrule
\multirow{4}{*}{FC2-500} & DO 0.0 & 54.75$\pm$0.26\% & 53.41$\pm$0.35\% \\
& DO 0.1 & 55.35$\pm$0.38\% & 54.06$\pm$0.46\% \\
& DO 0.25 & 55.81$\pm$0.37\% & 54.57$\pm$0.33\% \\
& DA & 53.85$\pm$0.34\% & 52.54$\pm$0.34\% \\
\midrule
\multirow{4}{*}{FC2-1000} & DO 0.0 & 53.78$\pm$0.24\% & 52.68$\pm$0.25\% \\
& DO 0.1 & 53.87$\pm$0.49\% & 52.45$\pm$0.15\% \\
& DO 0.25 & 54.87$\pm$0.43\% & 53.29$\pm$0.31\% \\
& DA & 52.59$\pm$0.20\% & 51.27$\pm$0.21\% \\
\midrule
\multirow{4}{*}{\shortstack{CONV\\(random)}} & DO 0.0 & 33.08$\pm$0.31\% & 32.65$\pm$0.38\% \\
& DO 0.1 & 33.04$\pm$0.42\% & 32.57$\pm$0.34\% \\
& DO 0.25 & 34.71$\pm$0.37\% & 33.90$\pm$0.53\% \\
& DA & 31.52$\pm$0.25\% & 31.04$\pm$0.45\% \\
\midrule
\multirow{4}{*}{\shortstack{CONV\\(trained)}} & DO 0.0 & 38.69$\pm$0.78\% & 35.82$\pm$0.59\% \\
& DO 0.1 & 39.23$\pm$0.82\% & 35.17$\pm$0.91\% \\
& DO 0.25 & 40.08$\pm$1.03\% & 35.51$\pm$0.61\% \\
& DA & 38.43$\pm$0.86\% & 34.39$\pm$0.64\% \\
\bottomrule
\end{tabular}
\end{table}

\vspace*{\fill}

\newpage

\subsection{Detailed proof of alignment between the BP and DRTP modulatory signals} \label{sec_supplementary_proof}

This full version of the alignment proof between the BP and DRTP modulatory signals is derived for a neural network composed of linear hidden layers~(Figure~\textbf{4}) and a single training example ($x$,$\:c^*$), where $x$ is the input data sample and $c^*$ the label. The $C$-dimensional target vector $y^*$ corresponds to the one-hot encoding of $c^*$, where $C$ is the number of classes. Our developments build on the alignment proof of~\cite{Lillicrap16}, which demonstrates that the FA and BP modulatory signals are within 90$^\circ$ of each other in the case of a single linear hidden layer, a linear output layer and a mean squared error loss. In the framework of classification problems, we extend this proof for the case of DRTP and to an arbitrary number of linear hidden layers, a nonlinear output layer of sigmoid/softmax units and a binary/categorical cross-entropy loss.

\paragraph{\textit{Network dynamics.}} The output of the $k$-th linear hidden layer is given by
\begin{equation*}
	y_k = z_k = W_k y_{k-1} \quad \text{for} \: k \in [1,K-1],
\end{equation*}
where $K$ is the number of layers and $y_0 = x$ is the input vector. Note that the bias vector $b_k$ is omitted without loss of generality. The output layer is described by
\begin{align*}	
	\begin{aligned}
		z_K &= W_K y_{K-1},\\
		y_K &= \sigma\left(z_K\right),
	\end{aligned}
\end{align*}
where $\sigma(\cdot)$ is either the sigmoid or the softmax activation function. The loss function $J(\cdot)$ is either the binary cross-entropy (BCE) loss for sigmoid output units or the categorical cross-entropy (CCE) loss for softmax output units, computed over the $C$ output classes:
\begin{align*}
	J_{\text{BCE}}(y_K, y^*) &= -\frac{1}{C} \sum_{c=1}^{C} \Big( y_c^* \log\left(y_{Kc}\right) + (1-y_c^*) \log\left(1-y_{Kc}\right) \Big),\\
	J_{\text{CCE}}(y_K, y^*) &= -\frac{1}{C} \sum_{c=1}^{C} \Big( y_c^* \log\left(y_{Kc}\right) \Big).
\end{align*}

The network is trained with stochastic gradient descent. In the output layer, the weight updates of both BP and DRTP follow
\begin{equation*}
	W_{K,ji} \leftarrow W_{K,ji} - \eta \sum_{l=1}^C \parderiv{J}{z_{Kl}} \parderiv{z_{Kl}}{W_{K,ji}}
\end{equation*}
where $i,j \in \mathbb{N}$ are indices corresponding respectively to the columns and rows of the output layer weight matrix. For both sigmoid and softmax output units, the factors in this update can be computed as
\begin{align*}
	\parderiv{J}{z_{Kl}} &= \sum_{c=1}^{C} \parderiv{J}{y_{Kc}} \parderiv{y_{Kc}}{z_{Kl}},\\
	\parderiv{z_{Kl}}{W_{K,ji}} &=
	\left\lbrace
		\begin{alignedat}{2}
		&y_{K-1,i} &&\quad \text{if}~~ j = l,\\
		&0 &&\quad \text{otherwise.}
		\end{alignedat}
	\right.
\end{align*}

For sigmoid output units, the factors in the partial derivative $\parderiv{J}{z_{Kl}}$ can be computed as
\begin{align*}
	\parderiv{J_{\text{BCE}}}{y_{Kc}} &= 
	\left\lbrace
		\begin{alignedat}{2}
		&-\frac{1}{C} \frac{1}{y_{Kc}} &&\quad \text{if}~~ c = c^*,\\
		&-\frac{1}{C} \frac{-1}{\left(1-y_{Kc}\right)} &&\quad \text{otherwise,}
		\end{alignedat}
	\right.\\
	\parderiv{y_{Kc}}{z_{Kl}} &= 
	\left\lbrace
		\begin{alignedat}{2}
		&y_{Kc} \left(1-y_{Kc}\right) &&\quad \text{if}~~ l = c,\\
		&0 &&\quad \text{otherwise,}
		\end{alignedat}
	\right.
\end{align*}
while for softmax output units, these factors can be computed as
\begin{align*}
	\parderiv{J_{\text{CCE}}}{y_{Kc}} &= 
	\left\lbrace
		\begin{alignedat}{2}
		&-\frac{1}{C} \frac{1}{y_{Kc}} &&\quad \text{if}~~ c = c^*,\\
		&0 &&\quad \text{otherwise,}
		\end{alignedat}
	\right.\\
	\parderiv{y_{Kc}}{z_{Kl}} &= 
	\left\lbrace
		\begin{alignedat}{2}
		&y_{Kc} \left(1-y_{Kc}\right) &&\quad \text{if}~~ l = c,\\
		&-y_{Kc} \: y_{Kl} &&\quad \text{otherwise.}
		\end{alignedat}
	\right.
\end{align*}

In both cases, it results that
\begin{equation*}
	\parderiv{J}{z_{Kl}} = 
	\left\lbrace
		\begin{alignedat}{2}
		&-\frac{1}{C} \left(1-y_{Kl}\right) &&\quad \text{if}~~ l = c^*,\\
		&-\frac{1}{C} \left(-y_{Kl}\right) &&\quad \text{otherwise,}
		\end{alignedat}
	\right.
\end{equation*}
which is equivalent to
\begin{align*}
	\parderiv{J}{z_K} &= -\frac{1}{C} \left(y^*-y_K\right) = -\frac{e}{C},
\end{align*}
where $e$ is the error vector. Therefore, the weight updates in the output layer can be rewritten as
\begin{equation*}
	W_K \leftarrow W_K + \frac{\eta}{C} e y_{K-1}^T.
\end{equation*}

In the hidden layers, the weight updates follow
\begin{equation*}
	W_k \leftarrow W_k - \eta \delta y_k y_{k-1}^T.
\end{equation*}

On the one hand, if the training relies on the BP algorithm, the modulatory signals $\delta z_k$, which are equivalent to the estimated loss gradients $\delta y_k$ in the linear case, correspond to the loss function gradient:
\begin{equation*}
	\delta y_k = \delta z_k = \parderiv{J}{y_k} = -\frac{1}{C} \left( \prod_{i=k+1}^K W_i^T \right) e.
\end{equation*}
On the other hand, if the DRTP algorithm is used, the modulatory signals are projections of the one-hot-encoded target vector $y^*$ through fixed random connectivity matrices $B_k$:
\begin{equation*}
	\delta y_k = \delta z_k = B_k^T y^*.
\end{equation*}

In order to provide learning, the modulatory signals prescribed by BP and DRTP must be within 90$^\circ$ of each other, i.e.~their dot product must be positive:
\begin{equation*}
	-e^T \left( \prod_{i=k+1}^K W_i^T \right)^T B_k^T y^* > 0.
\end{equation*}

\paragraph{\textit{Lemma.}} In the case of zero-initialized weights, i.e.~$W_k^0 = 0$ for $k \in [1,K]$, $k \in \mathbb{N}$, and hence of zero-initialized hidden layer outputs, i.e.~$y_k^0=0$ for $k \in [1,K-1]$ and $z_K^0 = 0$, considering a DRTP-based training performed recursively with a single element of the training set $(x,c^*)$ and $y^*$ denoting the one-hot encoding of $c^*$, at every discrete update step $t$, there are non-negative scalars $s_{y_k}^t$ and $s_{W_k}^t$ for $k \in [1,K-1]$ and a $C$-dimensional vector $s_{W_K}^t$ such that
\begin{align*}
	\begin{alignedat}{4}
		&y_k^t &&= -&&s_{y_k}^t \left( B_k^T y^* \right) \qquad &&\text{for} \qquad k \in [1,K-1]\\
		&W_1^t &&= -&&s_{W_1}^t \left( B_1^T y^* \right) x^T &&\\
		&W_k^t &&=  &&s_{W_k}^t \left( B_k^T y^* \right) \left( B_{k-1}^T y^* \right)^T \qquad &&\text{for} \qquad k \in [2,K-1]\\
		&W_K^t &&= -&&s_{W_K}^t \left( B_{K-1}^T y^* \right)^T. &&
	\end{alignedat}
\end{align*}

\textit{Proof.} The lemma is proven by induction.

For $t=0$, the conditions required to satisfy the lemma are trivially met by choosing $s_{y_k}^0,s_{W_k}^0 = 0$ for $k \in [1,K-1]$, and $s_{W_K}^0$ as a zero vector, given that $y_k^0=0$ for $k \in [1,K-1]$ and $W_k^0 = 0$ for $k \in [1,K]$.

For $t>0$, considering that the conditions are satisfied at a given discrete update step $t$, it must be shown that they still hold at the next discrete update step $t+1$. In the hidden layers, the weights are updated using the modulatory signals prescribed by DRTP. For the first hidden layer, we have
\begin{align*}
	W_1^{t+1} &= W_1^t - \eta B_1^T y^* x^T\\
	&= -s_{W_1}^t \left( B_1^T y^* \right) x^T - \eta \left( B_1^T y^* \right) x^T\\
	s_{W_1}^{t+1} &= s_{W_1}^t + \eta ~= s_{W_1}^t + \Delta s_{W_1}^t
\end{align*}
and for subsequent hidden layers, i.e.~for $k \in [2,K-1]$, we have
\begin{align*}
	W_k^{t+1} &= W_k^t - \eta B_k^T y^* y_{k-1}^{tT}\\
	&= s_{W_k}^t \left( B_k^T y^* \right) \left( B_{k-1}^T y^* \right)^T + \eta s_{y_{k-1}}^t \left( B_k^T y^* \right) \left( B_{k-1}^T y^* \right)^T\\
	s_{W_k}^{t+1} &= s_{W_k}^t + \eta s_{y_{k-1}}^t ~= s_{W_k}^t + \Delta s_{W_k}^t.
\end{align*}
The weights in the output layer are updated according to the loss function gradient, thus leading to
\begin{align*}
	W_K^{t+1} &= W_K^t + \frac{\eta}{C} \left(y^*-y_K^t\right) y_{K-1}^{tT}\\
	&= W_K^t - \frac{\eta}{C} \left(y^*-y_K^t\right) s_{y_{K-1}}^t \left( B_{K-1}^T y^* \right)^T\\
	&= -s_{W_K}^t \left( B_{K-1}^T y^* \right)^T - \frac{\eta s_{y_{K-1}}^t}{C} \left(y^*-y_K^t\right) \left( B_{K-1}^T y^* \right)^T\\
	s_{W_K}^{t+1} &= s_{W_K}^t + \frac{\eta s_{y_{K-1}}^t}{C} \left(y^*-y_K^t\right).
\end{align*}
	
The output of the first hidden layer is
\begin{align*}
	y_1^{t+1} &= W_1^{t+1} x\\
	&= \left( W_1^{t} - \eta B_1^T y^* x^T \right) x\\
	&= \underbrace{W_1^t x}_{y_1^t} - \eta x^T x \left(B_1^T y^* \right)\\
	&= -s_{y_1}^t \left(B_1^T y^* \right) - \eta \norm{x}^2 \left(B_1^T y^* \right)\\
	s_{y_1}^{t+1} &= s_{y_1}^t + \eta \norm{x}^2 ~= s_{y_1}^t + \Delta s_{y_1}^t
\end{align*}
and the output of the $k$-th hidden layer for $k \in [2,K-1]$ is given by
\begin{align*}
	y_k^{t+1} &= W_k^{t+1} y_{k-1}^{t+1}\\
	&= -s_{W_k}^{t+1} \left( B_k^T y^* \right) \left( B_{k-1}^T y^* \right)^T s_{y_{k-1}}^{t+1} \left( B_{k-1}^T y^* \right)\\
	&=- \left( s_{W_k}^t + \eta s_{y_{k-1}}^t \right) \left( s_{y_{k-1}}^t + \Delta s_{y_{k-1}}^t \right) \norm{B_{k-1}^T y^*}^2 \left( B_k^T y^* \right)\\
	&= -\underbrace{s_{W_k}^t s_{y_{k-1}}^t \norm{B_{k-1}^T y^*}^2}_{s_{y_k}^t} \left( B_k^T y^* \right) - \left( s_{W_k}^t \Delta s_{y_{k-1}}^t  + \eta s_{y_{k-1}}^t \left(s_{y_{k-1}}^t + \Delta s_{y_{k-1}}^t \right) \right) \norm{B_{k-1}^T y^*}^2 \left( B_k^T y^* \right)\\
	s_{y_k}^{t+1} &= s_{y_k}^t + \left( s_{W_k}^t \Delta s_{y_{k-1}}^t  + \eta s_{y_{k-1}}^t \left(s_{y_{k-1}}^t + \Delta s_{y_{k-1}}^t \right) \right) \norm{B_{k-1}^T y^*}^2 ~= s_{y_k}^t + \Delta s_{y_k}^t.
\end{align*}

The coefficients $s_{W_1}^t$ and $s_{y_1}^t$ are updated with strictly positive quantities $\Delta s_{W_1}^t$ and $\Delta s_{y_1}^t$ at each update step $t$ and are thus strictly positive for $t>0$. Furthermore, the coefficients $s_{W_k}^t$ and $s_{y_k}^t$ are updated based on the coefficients of the previous layer and will therefore be strictly positive for $k \in [1,K-1]$.
\qed

\paragraph{\textit{Theorem.}} Under the same conditions as in the lemma and for the linear-hidden-layer network dynamics described above, the $k$-th layer modulatory signals prescribed by DRTP are always a negative scalar multiple of the Moore-Penrose pseudo-inverse of the product of forward matrices of layers $k+1$ to $K$, located in the feedback pathway between the output layer and the $k$-th hidden layer, multiplied by the error. That is, for $k \in [1,K-1]$ and $t>0$,
\begin{equation*}
	- \frac{1}{s_k^t} \left( \prod_{i=K}^{k+1} W_i^t \right)^+ e = B_k^T y^* \quad \text{with} \quad s_k^t > 0.
\end{equation*}

\textit{Proof.} When replacing the forward weights $W_i^t$ by the expressions given in the lemma, the above equality becomes
\begin{align*}
	\left[ \left(\prod_{i=K-1}^{k+1} s_{W_i}^t \right) s_{W_K}^t \left(\prod_{i=K-1}^{k+1} \norm{B_i^T y^*}^2 \right) \left( B_k^T y^* \right)^T \right]^+ \left(y^*-y_K^t\right) &= s_k^t B_k^T y^*\\
	\left(\prod_{i=k+1}^{K-1} s_{W_i}^t \right)^{-1} \left(\prod_{i=k+1}^{K-1} \norm{B_i^T y^*}^2 \right)^{-1} \left[ s_{W_K}^t \left( B_k^T y^* \right)^T \right]^+ \left(y^*-y_K^t\right) &= s_k^t B_k^T y^*\\
	\left(\prod_{i=k+1}^{K-1} s_{W_i}^t \right)^{-1} \left(\prod_{i=k+1}^{K-1} \norm{B_i^T y^*}^2 \right)^{-1} \left( B_k^T y^* \right)^{T+} s_{W_K}^{t+} \left(y^*-y_K^t\right) &= s_k^t B_k^T y^*\\
	\left(\prod_{i=k+1}^{K-1} s_{W_i}^t \right)^{-1} \left(\prod_{i=k+1}^{K-1} \norm{B_i^T y^*}^2 \right)^{-1} \underbrace{\norm{B_k^T y^*}^{-2} \left(B_k^T y^*\right)}_{\left( B_k^T y^* \right)^{T+}} \underbrace{\norm{s_{W_K}^t}^{-2} s_{W_K}^{tT}}_{s_{W_K}^{t+}} \left(y^*-y_K^t\right) &= s_k^t B_k^T y^*\\
	\left(\prod_{i=k+1}^{K-1} s_{W_i}^t \right)^{-1} \left(\prod_{i=k}^{K-1} \norm{B_i^T y^*}^2 \right)^{-1} \norm{s_{W_K}^t}^{-2} s_{W_K}^{tT} \left(y^*-y_K^t\right) \left(B_k^T y^*\right) &= s_k^t \left(B_k^T y^*\right).\\
\end{align*}
By identification, it is found that
\begin{equation*}
	s_k^t = \dfrac{s_{W_K}^{tT}\left(y^*-y_K^t\right)}{\left(\prod_{i=k+1}^{K-1} s_{W_i}^t \right) \left(\prod_{i=k}^{K-1} \norm{B_i^T y^*}^2 \right) \norm{s_{W_K}^t}^2} ~.
\end{equation*}

From the lemma proof, the update formula for the vector $s_{W_K}^t$ is given by
\begin{equation*}
	s_{W_K}^{t+1} = s_{W_K}^t + \frac{\eta s_{y_{K-1}}^t}{C} \left(y^*-y_K^t\right),
\end{equation*}
where $\eta$, $C$ and $s_{y_{K-1}}^t$ are positive scalars. In the framework of classification problems where outputs are strictly bounded between 0 and 1, for any example ($x$,$\:c^*$) in the training set, the error vector $e = \left(y^*-y_K^t\right)$ has a single strictly positive entry $(1-y_{Kc})$ at the class label index $c=c^*$, all the other entries $-y_{Kc}$ with $c \neq c^*$ being strictly negative. This sign information is constant as the network is trained with a single training example. Given that $s_{W_K}^0 = 0$ from zero-weight initialization and that $s_{W_K}^t$ is updated in the same direction as $e$, we have at every discrete update step $t$
\begin{equation*}
	\text{sign} \left(s_{W_K}^t\right) = \text{sign} \left(y^*-y_K^t\right),
\end{equation*}

and thus
\begin{equation*}
	s_{W_K}^{tT} \left(y^*-y_K^t\right) > 0.
\end{equation*}

Therefore, the scalars $s_k^t$ are strictly positive for $t>0$. \qed

\paragraph{\textit{Alignment.}} In the framework of classification problems, as the coefficients $s_k^t$ are strictly positive scalars for $t>0$, it results from the theorem that the dot product between the BP and DRTP modulatory signals is strictly positive,~i.e.
\begin{align*}
	\begin{aligned}
	- e^T \left( \prod_{i=k+1}^K W_i^T \right)^T \left( B_k^T y^*\right) &> 0\\
	e^T \underbrace{\left( \prod_{i=k+1}^K W_i^T \right)^T  \left( \prod_{i=K}^{k+1} W_i \right)^+}_{I} \frac{e}{s_k^t} &> 0\\
	\frac{e^Te}{s_k^t} &> 0.
	\end{aligned}
\end{align*}

The BP and DRTP modulatory signals are thus within 90$^\circ$ of each other. \qed

\end{document}